\documentclass[11pt]{article}
\usepackage[margin=1in]{geometry}

\usepackage{amsmath, amsthm, amssymb, amsfonts}
\usepackage{algpseudocode}
\usepackage{algorithm}
\usepackage{lineno}

\algtext*{EndIf} 
\algtext*{EndWhile} 
\algtext*{EndFor} 
\algtext*{EndFunction} 
\algtext*{EndProcedure} 

\usepackage{graphicx}
\usepackage{booktabs}
\usepackage{fvextra}
\usepackage{bm}

\newcommand{\indep}{\perp \!\!\! \perp}

\usepackage{xcolor}

\usepackage{hyperref}
\usepackage[
backend=biber,
style=authoryear-comp,
]{biblatex}
\title{Belief as a Predictive Abstraction for Language Models}
\title{Beliefs and Behavior in Language Models}
\author{
	Alex Smolin\\
    Department of Economics\\
Toulouse School of Economics\\
alex.smolin@tse-fr.eu 
	\and 
Bryan Wilder\\
	Machine Learning Department\\ Carnegie Mellon University\\
	bwilder@cmu.edu
 }
\date{}
\usepackage{titling}
\begin{document}
\maketitle

\begin{abstract}
    There is significant uncertainty about whether abstractions like \textit{beliefs} or \textit{desires} usefully describe the behavior of large language models (LLMs). In addition to the inherent scientific interest of this question, these latent quantities are often invoked to explain the behavior of LLMs to users or to define and evaluate harmful behaviors which are relative to intent. Nevertheless, we currently lack a means to systematically test whether concepts like ``belief" are well-applied to LLMs, and hence whether they are likely to be fruitful ingredients of attempts to align models with human interests. We propose an approach for empirically studying such questions, asking whether a single latent variable inferred from the LLMs' outputs -- interpreted as a degree of belief -- allows an observer to make interpretable predictions of how the LLMs' will respond to new prompts. We find that highly capable models are usefully described as holding beliefs and that, generally, the predictability of model outputs based on an inferred latent belief tracks overall trends in model capability. Building on these findings, we provide empirical strategies to study how beliefs in LLMs can be measured, the extent to which LLMs comply with instructed decision rules or payoffs, and how beliefs evolve within individual instances of an LLM over the course of reasoning.       
\end{abstract}

\section*{Introduction}

There is significant debate across computer science, cognitive science, philosophy, and economics about whether it is legitimate to attribute cognitive states like holding beliefs, desires and goals to LLMs \parencite{cappelen2025going,ibrahim2026thinking,herrmann2024standards,goldstein2024does,mitchell2023debate,shanahan2024talking}. One prominent philosophical school of thought is an interpretationist viewpoint, summarized by \textcite{goldstein2025does}: “a system has particular beliefs and desires if and only if its actions are predicted sufficiently well by the hypothesis that it is rational and has those beliefs and desires.” Setting aside whether the interpretationist standpoint is fundamentally correct about the nature of belief, understanding whether actions taken by LLMs can be predicted by abstractions like beliefs and desires, and if so, developing strategies to elicit such quantities, has important implications for AI alignment. For users, such characteristics provide a route to make decision-making legible. For evaluators, these ingredients are presumed in defining and measuring behaviors like deception that are inherently relative to beliefs. For developers, such characteristics could serve ingredients in training processes which shape the underlying mechanisms producing behavior directly, instead of supervising only outputs, as suggested by approaches like  “character training” \parencite{anthropic2024character}.

The field currently lacks a means to systematically test the predictive utility of attributing beliefs to LLMs -- whether this attribution allows an observer to successfully predict what the model will do in new circumstances. \textit{Our goal is to provide such a framework for the case of problems where an agent must take actions while facing an unknown, binary state of the world}. While such decision problems are certainly a very limited subset of the purposes for which language models are employed, focusing on problems of this form allows us to provide a mathematically precise evaluation, which we argue is valuable in untangling competing perspectives and offering rigorous metrics of success.

In particular, answering whether attributing beliefs to LLMs is useful requires a formalization of the consequences of holding a belief. For example, some have argued that LLMs lack internal coherence: that they can be led to take contradictory actions based on different framings of a problem \parencite{binz2023using,echterhoff2024cognitive}, that their stated beliefs and actions are inconsistent \parencite{pal2025knowing}, or that their beliefs, when expressed in probabilistic terms, do not obey the relevant axiomatic properties \parencite{zhu2024incoherent,chen2026llms}. On the one hand, these failures are problematic because they call into question the “rational” component of the definition above. On the other hand, all of these various contradictions are present in humans: the process of converting our fuzzy beliefs into numerically precise verbalizations or actions is imperfect, susceptible to influence by the elicitation process itself, and certainly not consistent with probability axioms. How then should we decide whether there is a sufficiently coherent structure to an agent’s actions so that beliefs are usefully predictive of behavior? And how can those latent beliefs be measured?

We propose an approach that treats a belief about an unknown as a latent variable which may be reflected only noisily in any particular output from an agent. In order to fit this latent variable model, we query the output of an LLM to many different prompts on many cases drawn from some distribution. The model then infers, using the LLMs’ responses to the different prompts, the value of this latent state for each case. On the hypothesis that the LLM acts as if it holds beliefs, this latent variable should have structured and highly predictive relationships with each of its outputs: a single number (the value of the latent belief state) should suffice to reconstruct with reasonable fidelity what the LLM’s responses to the individual questions will be. On the other hand, if the LLM’s behavior is not internally coherent, such a reconstruction will be impossible. 

We implement this framework and use it to evaluate the behavioral coherence of a range of LLMs in several decision-making settings: a high-stakes healthcare setting (screening for chronic kidney disease), a strategic interaction under imperfect information (games of Werewolf), and a scientific task (forecasting the results of social science experiments). For the most capable models, we find that behavior is highly internally coherent: a single latent state suffices to reconstruct behavior across 16 different prompts asking for different kinds of outputs or decisions with high fidelity (AUCs above 0.90). We therefore conclude that, at least in such settings, ``belief" is a useful abstraction to predict the behavior of highly capable LLMs.  

We then leverage our framework to contribute to several additional questions about the nature of beliefs and actions in LLMs. 

First, how can latent beliefs in LLMs be measured?  Our framework allows us to assess the quality of specific prompts at serving as measurements of a latent belief, rather than treating the response to any single prompt as identical with the belief (as previous work often does). We find that verbalized probability judgments from models are relatively high-quality belief measurements, especially for highly capable models. 

Second, can models be described as internally coherent while falling prey to framing effects or giving outwardly incompatible responses? For example, if a model describes an event as happening with 60\% probability but rejects a bet on the event at 2:1 odds, is this evidence of a failure to hold coherent beliefs in the first place? We first show that the prevalence of such effects is substantially reduced in the most capable models. For less capable models which demonstrate outward incoherence, our results suggest that their responses to different decision framings can sometimes be interpreted as decoding a shared latent variable, but the numerical scale of that mapping may not be instruction-following or mutually consistent across framings.

Third, which entity should inferred beliefs should be associated with? For example, previous work has distinguished between hypotheses that beliefs should be associated with the base model, with an individual instance of the model (i.e., the base model + a particular prefix in the context window), or with intermediate entities like “the model while inhabiting a particular persona” \parencite{chalmers2025we,goldstein2025does,beckmann2026mind}. While we do not claim to decisively resolve these questions, our framework provides a way of making them precise by measuring the concordance in beliefs between these candidate entities. As an illustration, we “branch” the model after an initial chain-of-thought phase and measure the similarity between independently elicited beliefs for instances along a common branch. We find that models’ beliefs evolve meaningfully over the course of deliberation and so there will be circumstances where it is useful to distinguish between the beliefs held by individual instances as opposed to the base model.  

\section*{Methods}

\subsection*{Latent variable model}

We consider cases $(x, y)$ drawn from a distribution $Q$, where $x$ is an observable context and $y$ is an unknown state. For simplicity, $y$ is binary in our three empirical examples. The model is presented with $x$ along with additional instruction $c$, which consists of direction from the user about the decision problem to be solved. The model then outputs an action $a \in A$. As stated so far, the model may output entirely different responses given different instructions $c$, for the same context $x$. On the hypothesis that the model acts as if it holds a belief about $y$, though, its responses to different instructions $c_1, c_2, ..c_J$ will be related to one another. We model this dependence by introducing a latent variable $z \in [0,1]$, which we interpret as the degree of belief that the model holds about $y$: $z$ is higher if the model has a greater degree of belief that $y = 1$, and lower if the model has a greater degree of belief that $y = 0$. Formally, we posit the following probabilistic latent variable model:
\begin{align*}
&z \sim P(\eta)\\
&a_j | z \sim F_j(z), \,\,\,\text{independently for }j = 1...J
\end{align*}
where $P$ is a prior distribution parameterized by $\eta$ and $F_j$ controls the distribution of each output $a_j$ conditional on $z$. Without further structure, though, this formulation does not impose the semantics typically associated with the concept of a belief. Indeed, if the $F_j$ are sufficiently expressive, a single real number ($z$) can in principle encode arbitrary behavior. Accordingly, predictive and interpretational utility will require us to impose restrictions on the class of the $F_j$ which relate latent beliefs to actions. 

Because we focus on decision problems with binary state, guidance about these restrictions is available from decision theory and the discrete choice literature. \textcite{yamin2026agents} show that if $z$ is the belief of an agent which acts “approximately” as an expected utility maximizer, it should have two properties.  First, beliefs should be a sufficient statistic, containing all outcome-relevant information used by the agent in its decisions. Our model structure enforces a kind of internal sufficiency across multiple actions: since the $a_j$ are sampled independently conditional on $z$, $z$ must contain all latent information that might correlate the different responses from the model. We also empirically evaluate \textcite{yamin2026agents}'s sense of outcome sufficiency (that $y \perp a_j | z$, i.e., $z$ summarizes all information about $y$ used to produce $a_j$). Residual outcome dependence is typically small for capable models (Supplementary Information, Section~\ref{app:outcome-sufficiency}).

Second, choice probabilities should satisfy a property known as cyclic monotonicity. In the case of binary actions, this is simply that the probability of taking $a = 1$ should be monotone in the latent belief, corresponding to the interpretation that the agent prefers one action more when the state  is 1 and the other when the state is 0. We will be able to enforce this condition by making particular choices for the functional form of the $F_j$. Similarly, we may also wish to impose other common-sense interpretational constraints, like that when the instructed output is a numerical probability, the distribution over that probability should be monotone in the latent state (e.g., in the sense of stochastic dominance). If we impose such constraints on the form of the model, and it still predicts the LLM’s behavior well, then there is a more useful sense in which the LLM is describable as having beliefs since the latent belief predicts behavior through an interpretable set of preferences.

Concretely, in our main specification, we adopt an extremely simple functional form where for binary actions
\begin{align*}
    \text{logit} \Pr(a_j = 1|z) = a_j + b_j\cdot \text{logit}(z)
\end{align*}
with ordinal actions following a similar logit model (see Supplementary Information). When $c_j$ asks the model to output a probability, we use the specification
\begin{align*}
    \Pr(a_j|z) = \text{Beta}(z\phi, (1-z)\phi) 
\end{align*}
where $\phi$ is a parameter scaling the precision of probability outputs. This choice anchors the scale of the latent to correspond to a probability, while allowing stochasticity in the realization of the verbalized probability output conditional on $z$. These functional forms restrict the LLM to have choice probabilities which are monotone in z, and which for binary actions are interpretable as stochastic choices around the decision threshold $\operatorname{logit}^{-1}(-a_j/b_j)$. The prior distribution $P$ is then modeled via a density that is piecewise linear in log-space, parameterized by the values at 7 control points that are jointly inferred with the other parameters. We find that despite its simplicity, this model describes LLMs’ observed choices well.

In the Supplementary Information, we explore two ways of making the model more expressive. First, we allow $z$ to belong to $\mathbb{R}^2$ instead of $\mathbb{R}$, testing the hypothesis that LLM behavior is better-explained by a higher-dimensional latent variable (perhaps, beliefs or dispositions with respect to multiple quantities). Second, we allow the relationship between $z$ and outputs to be more expressive by allowing $F_j$ to use a nonlinear spline instead of a generalized linear model. Neither of these extensions provide a detectable improvement in predictive power and so we present the simpler model in which $z$ and the $F_j$ have clear interpretations.  

\subsection*{Inference and evaluation}

Given observed data of the form $\{(x_i, c_{ij}, a_{ij})\}$ across a set of cases $i$ and actions $j$, we can perform inference in the latent variable model to obtain a posterior distribution over the parameters $\theta = (\eta, \phi, \alpha_1, \beta_1, …, \alpha_J, \beta_J)$, as well as the unknown latent $z_i$. Our general process for each domain is to use a training set to produce a fixed maximum likelihood estimate for $\theta$, integrating out the unknown $z$ via Gauss–Legendre quadrature. Then for each input $x_i$ in a held-out test set, we infer a posterior over $z_i$ using the observed $\{(c_{ij}, a_{ij})\}$, conditional on the previously estimated $\hat{\theta}$. The test set prediction for each output is the predictive-posterior mean, integrating again over $z_i$ via quadrature. This asks whether an observer can predict what action the model will take in response to instruction $c_j$ after seeing the model’s responses to the other instructions and inferring based on these the model’s latent belief.  See the Supplementary Information for additional details.

\subsection*{Domains}

We study the behavior of LLMs in three domains, chosen to represent different contexts and forms of capability. 

First, a consequential medical setting: deciding whether a patient should be tested for chronic kidney disease based on baseline clinical and demographic information. This mimics a real public health decision made under uncertainty \parencite{stevens2024kdigo}. We use patient data from the United States National Health and Nutrition Examination Survey. 

Second, a social setting involving strategic interactions and incomplete information: games of Werewolf. In Werewolf, players are assigned hidden roles (one of which is the werewolf) and must, over multiple rounds, make deductions about these roles and persuade other players to coordinate on actions. Most players aim to reveal the werewolf, while the werewolf aims to eliminate the other players. We use the Kaggle Agent Arena Werewolf dataset \parencite{kaggle2026werewolf}, which consists of the full transcripts of games played between frontier LLMs. The model is given the transcript information available from the perspective of a focal non-werewolf player and asked to make decisions which depend on whether a given target player is the werewolf (e.g., whether, if it were the last round of the game, the LLM would vote to eliminate the target player). 

Third, a scientific domain where the LLM is asked to predict the results of social science survey experiments. We use the dataset of \textcite{ashokkumar2026large}, which collected  social science experiments both drawn from the NSF Time-sharing Experiments for the Social Sciences platform and from a curated set of mega-studies. In each case, the model’s input context is a full description of the stimulus presented to subjects in two experimental conditions and the target unknown is whether the mean value of an outcome variable will be higher in Condition A or Condition B. Based on this output, the LLM is asked to make decisions that depend on this unknown, e.g., which condition should be prioritized for a follow-up experiment. 

\subsection*{Action spaces}

In each domain, we query 16 different outputs for each of 500 cases (where an case is an input $x$, i.e., a patient, game, or pair of conditions). All responses use temperature=1. Each output corresponds to an instruction $c_j$ and action $a_j$, divided across three categories. First, assessments of the probability that $y = 1$ (where $y$ is an event specified for each domain). This includes requests for the model to output a probability as a number between 0 and 1, responses to a Likert scale, and multiple-choice questions picking out the range of [0,1] which contains $\Pr(y = 1|x)$. Second, a binary decision task specified for each domain, as in each example above (see the Supplementary Information for the exact prompts used). Third, parameterized decisions in which $c_j$ provide additional instruction to the model about the circumstances under which it should take a = 1. We consider several framings: ones in which $c_j$ specifies a target probability threshold above which the LLM should take $a = 1$, ones in which the LLM is given a target utility function specified in terms of the costs of false positives vs false negatives, and ones in which the LLM is told that $a = 1$ consists of making a bet with specified payoffs in the event that $y = 0$ vs 1. These three framings are constructed while varying the exact parameters to create paired instructions that would be equivalent to an expected utility maximizer across the three framings (e.g., a target threshold of 33\% is equivalent to false negatives being weighted 2x false positives, which is in turn equivalent to a bet at 2:1 odds). This allows us to distinguish behavior driven by the decision threshold implied by a given set of instructions from the effect of changes in the framing itself. 

\section*{Results}

\subsection*{Prediction}
\begin{figure}
    \centering
    \includegraphics[width=0.95\linewidth]{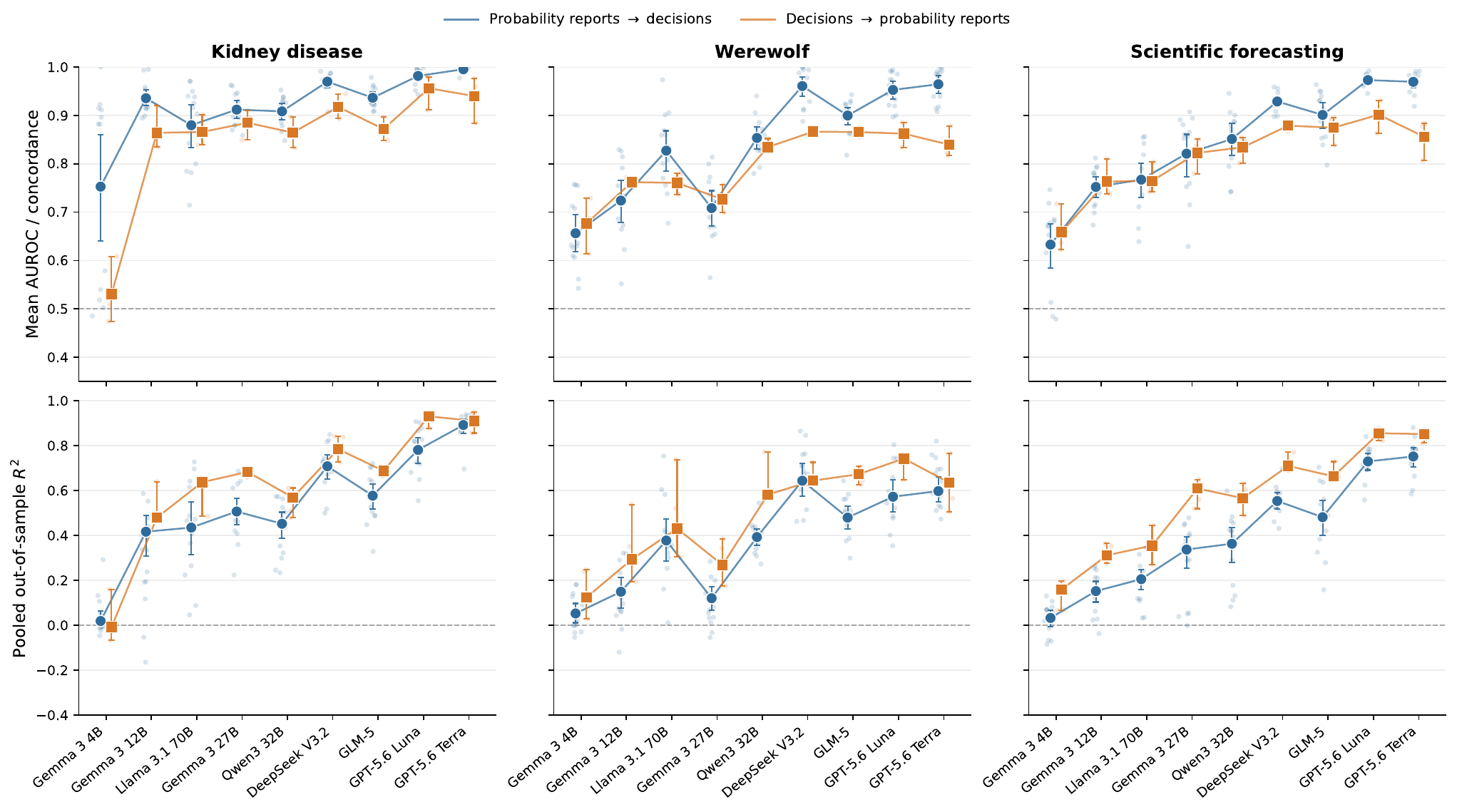}
    \caption{Out of sample performance at predicting a held-out output from the LLM. Top: performance assessed in terms of concordance index. Bottom: performance assessed in terms of $R^2$. Dark dots show the average performance across outputs for each model, while light dots show performance for each of the 16 individual outputs.}
    \label{fig:loo-prediction}
\end{figure}

We start by evaluating the extent to which the actions taken by LLMs are predictable via a single inferred latent belief. We hold out different sets of actions $a_j$ at test time, asking whether a belief that is inferred using observations of only a subset of the model's responses allows prediction of the remaining ones. In our main analysis, presented in Figure \ref{fig:loo-prediction}, we show results for two held-out action sets. In the first, all actions where $c_j$ asks the model to make a decision are held out, and predicted using responses to probability assessment questions. In the second, decisions are instead observed and used to predict the probability assessment responses. We evaluate predictive performance for a range of LLMs at different levels of capability using two metrics.  First, the concordance index, which generalizes AUROC from binary prediction to arbitrary outputs: it measures the fraction of cases in which the true targets are correctly ranked by the predictions. Second, the $R^2$ for predicting each output. Models are ordered by scores on the Artificial Analyses capabilities index.

\begin{figure}
    \centering
    \includegraphics[width=0.95\linewidth]{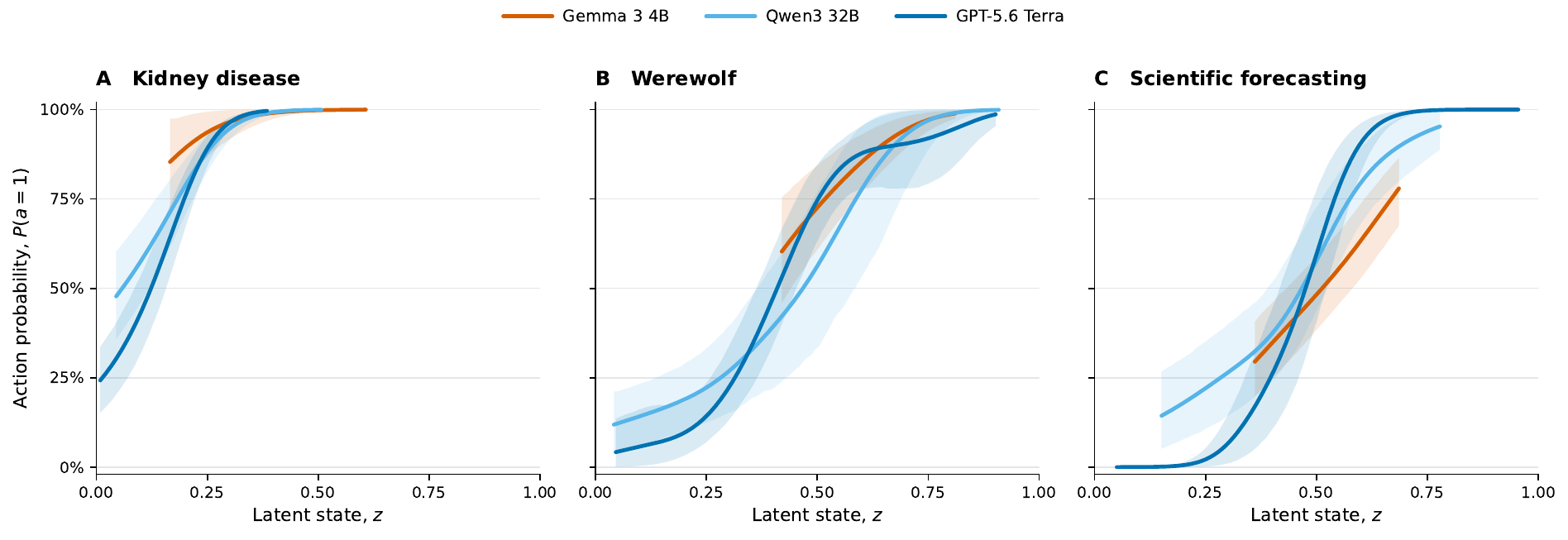}
    \caption{Visualized choice probabilities for taking $a = 1$ in the main decision task for each domain for three example models. The line shown is a Nadaraya-Watson kernel estimate of $\Pr(a = 1|z)$, with the shaded region giving a 95\% bootstrap confidence band.}
    \label{fig:decision-curves}
\end{figure}

We find that highly capable LLMs take actions across all three domains that are well-predicted by the latent variable model, with concordance indices in the range 0.8-0.95 and $R^2$ values in the range of 60-90\%. Less capable LLMs score much less highly, and the overall trend in performance aligns with a general gradient of model capabilities. Figure \ref{fig:decision-curves} shows examples of the empirical choice probabilities of individual models as a function of their inferred latent belief. We observe that less capable models have monotonic but weak and noisy relationships, while strong models move progressively closer to acting based on a threshold in the latent belief space.

We then evaluate stronger senses of generalizations for behavior prediction. So far, we have evaluated predictions where a given $a_j$ is withheld at test time, but the $F_j$ used to predict $a_j$ from $z$ was fit using supervised data. That is, we have evaluated whether an observer can predict an unseen action in a new case, but of a kind of action for which they have previously seen examples. Next, we evaluate whether inferred beliefs allow predictions that generalize zero-shot in two senses: to classes of instructions $c_j$ that were never observed during training, and to unseen domains. The former requires defining a class of $F$ to map from $z$ to predictions which can be applied zero-shot, and so we focus in this analysis on predicting outputs that belong to parameterized families of decisions (e.g., prompting the model with specified decision thresholds or betting odds). For these outputs, we parameterize $F_j$ as 
\begin{align}
\operatorname{logit}\Pr(a_j = 1|z)
&=
\alpha
+
\beta\left[
\operatorname{logit}(z_i)
-
\operatorname{logit}(\tau_j)
\right]
\end{align}
where $\tau_j$ is the decision threshold implied by instruction $c_j$ and the parameters $\alpha$ and $\beta$ are now shared across framings $j$. In across-framing generalization, the latent variable model is trained using data from two of the three framings and evaluated on the third. In across-domain generalization, the latent variable model is trained on two domains (but with all framings) and tested on the third. Finally, we evaluate combined generalization by training the framing-agnostic model above on two domains and evaluating on a third out-of-sample domain.

\begin{figure}
    \centering
    \includegraphics[width=0.95\linewidth]{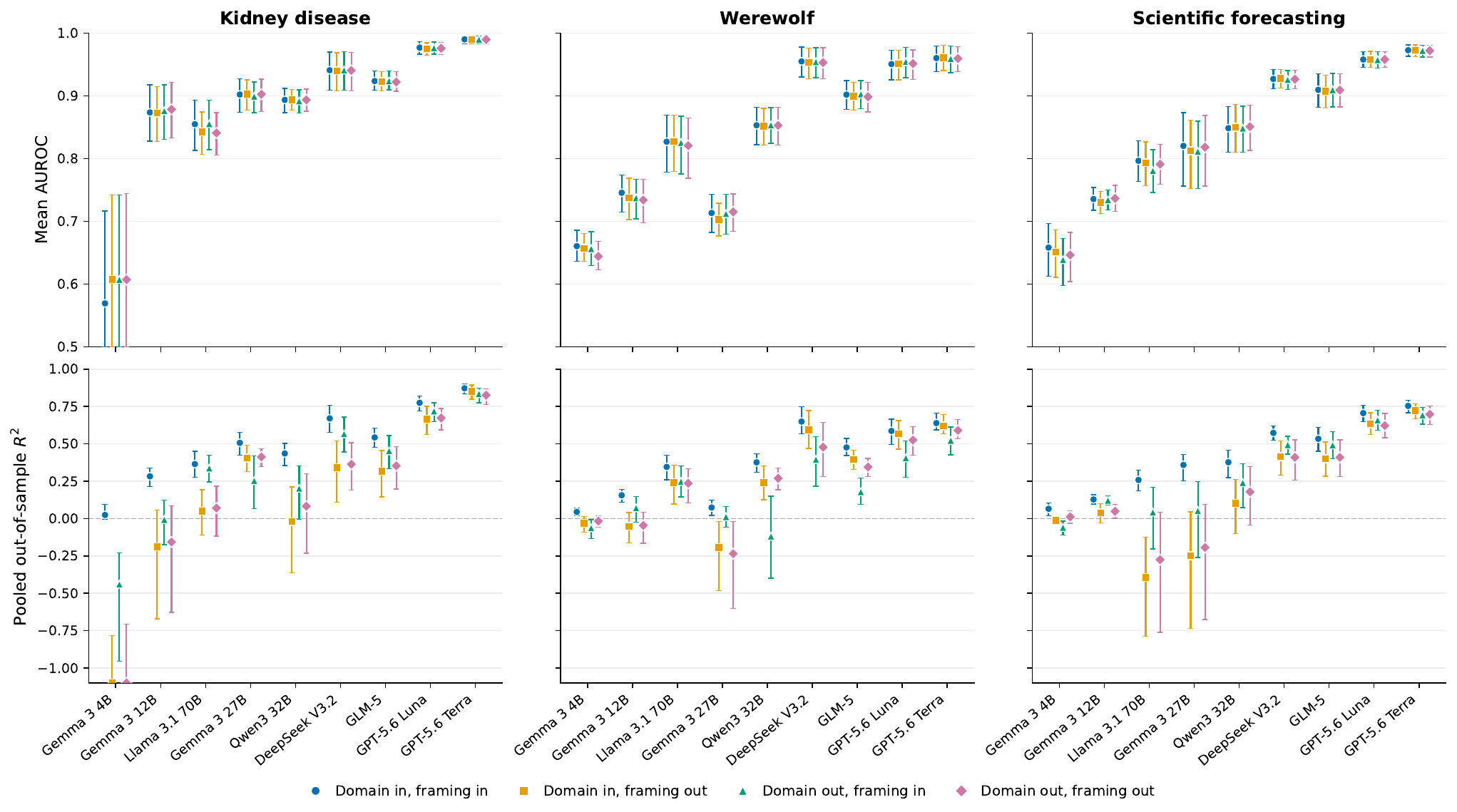}
    \caption{Out of domain and out of framing generalization in predicting LLM outputs. Top: performance assessed in terms of area under the ROC curve (AUROC). We refer to the metric as AUROC instead of concordance index because all prediction targets in this analysis are binary. Bottom: performance assessed in terms of pooled $R^2$.}
    \label{fig:out-of-domain}
\end{figure}

Figure \ref{fig:out-of-domain} shows the results. Matching our earlier analysis, we consider two settings: one where $z$ is inferred using the (held-in) decision outputs, and one where it is inferred using the probability-related outputs. Figure \ref{fig:out-of-domain} shows the first case, with the second in the Supplementary Information. We find that for the most capable models, predictions that generalize zero-shot across framings, domains, or both simultaneously remain highly accurate, with AUCs close to 1 and pooled $R^2$ values in the range of approximately 0.6 to 0.8. However, for less-capable models, such predictions are significantly less accurate than the original setting. This indicates that more capable models have a greater degree of coherence across settings such that accurate behavior prediction requires less setting-specific information. 

\subsection*{Measuring the sufficiency of elicited probabilities}
\begin{figure}
    \centering
    \includegraphics[width=0.95\linewidth]{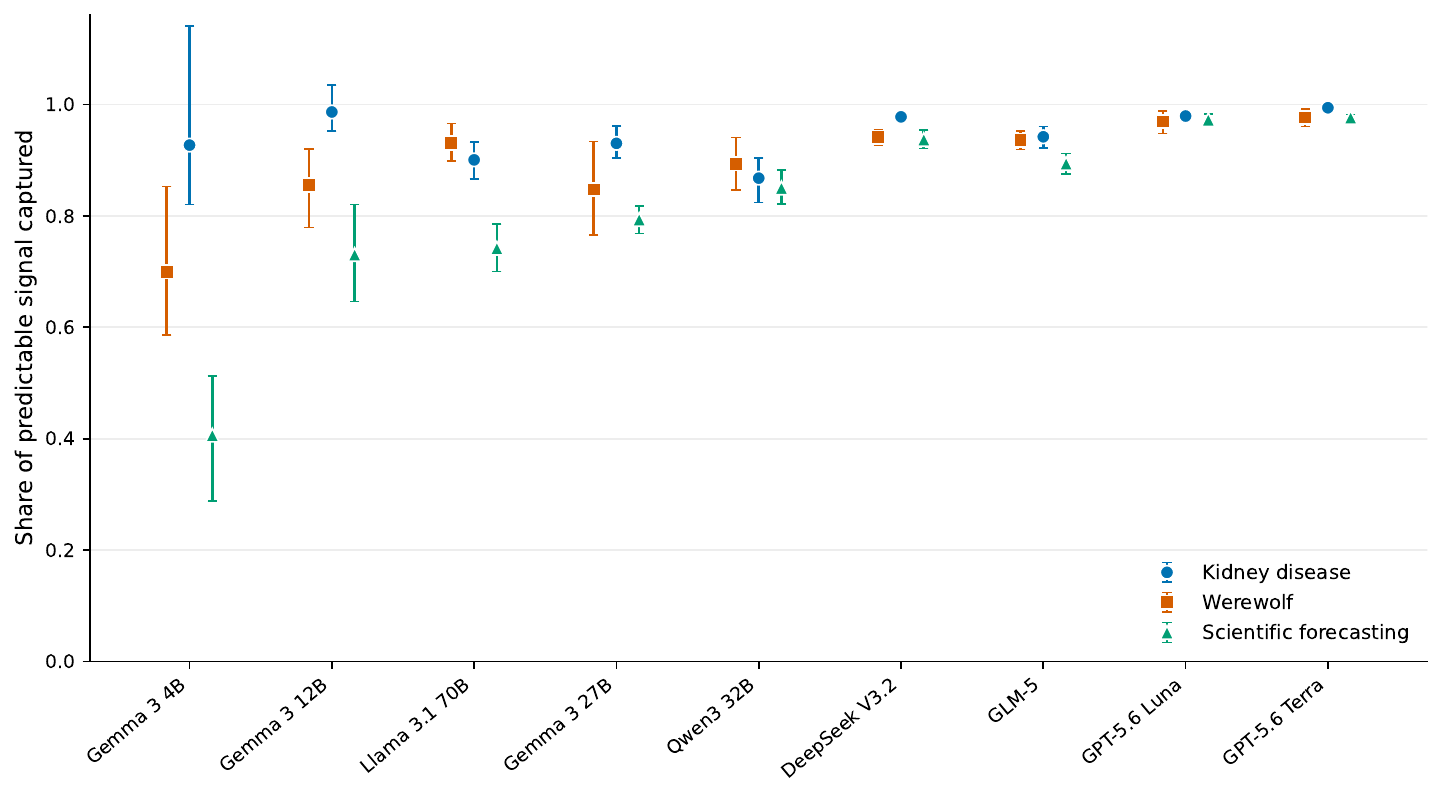}
    \caption{Fraction of held-out predictive performance captured by the numerical probability report. For each of the 15 non-probability targets, both predictions use the same fixed latent-variable model: one infers the latent posterior from the numerical probability alone and the other from all responses except the target. The y axis is the ratio of their macro-average excess concordance above chance, $(\overline c_{\mathrm{probability\ only}}-0.5)/(\overline c_{\mathrm{all\ other}}-0.5)$, using every target for which both concordances are defined. Figure \ref{fig:prob-sufficiency-r2} in the Supplementary Information shows the corresponding results with respect to $R^2$, which are highly similar.}
    \label{fig:prob-sufficiency}
\end{figure}

While our analysis so far uses a large number of responses from the model to infer its latent belief for each case, we may wonder whether this is really necessary. Certainly, it would be desirable for models to be able to give a good summary of their belief in response to a simple request. In Figure \ref{fig:prob-sufficiency}, we evaluate the fraction of predictive performance from the fully inferred latent belief that can be recovered via the answer to a single question which asks the model to verbalize a probability judgment about the focal event (i.e., to report an estimate of $\Pr(y=1|x)$). Both predictions use the same fitted latent-variable model: for each non-probability target, the comparison conditions the latent posterior either on all other outputs or only on the numerical probability report. We find that, especially for more capable models, single elicited probabilities are high-quality measurements of latent belief states: for the best models, the answer to this single prompt accounts for over 90\% of the predictive performance of the jointly inferred latent belief. For weak models, on the other hand, elicited probabilities are a significantly worse summary of their latent belief, which is in turn much less coherently related to their actions.

\subsection*{Inferred decision thresholds}
\begin{figure}
    \centering
    \includegraphics[width=0.95\linewidth]{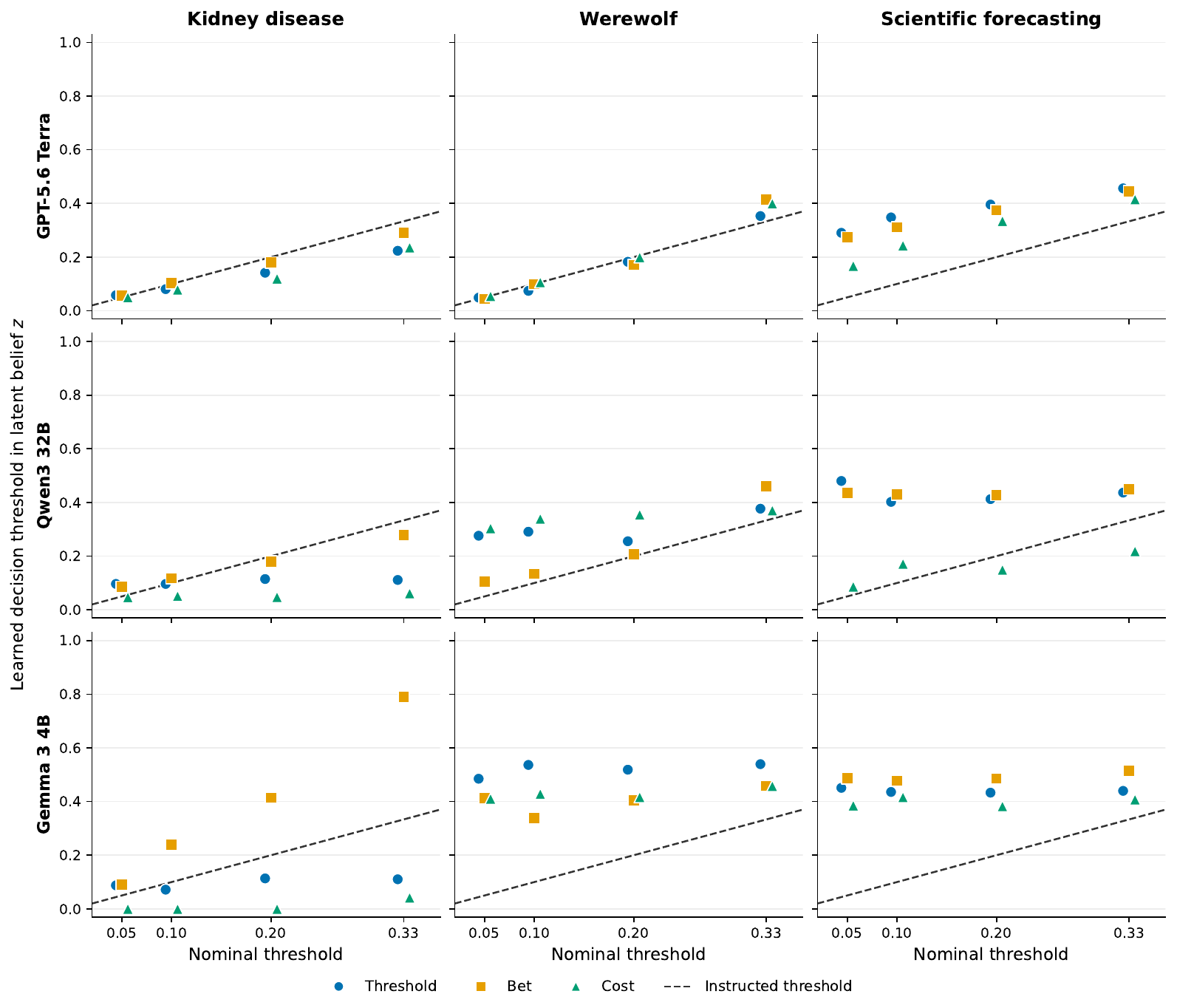}
    \caption{Maximum-likelihood decision thresholds on the probability-anchored latent-belief scale for each framing, as a function of the nominal threshold instructed in the prompt.}
    \label{fig:thresholds}
\end{figure}

While, for capable models, elicited probabilities are high-quality belief measurements in terms of information content, LLMs may not produce actions that are semantically consistent with both the elicited probability and the instructions expressed in the prompt $c_j$. To check this form of consistency, Figure \ref{fig:thresholds} shows the implied probability-space decision threshold for each domain and framing, visualized for three representative models (Gemma 3 4B, Qwen3 32B, and GPT-5.6 Terra). The three framings present payoff-identical decision problems in three ways: by specifying a target probability threshold for taking $a = 1$, by specifying costs for false positives and false negatives, and by presenting the problem as a bet with specified payoffs. For each framing, model, and domain, we show the estimate of $\operatorname{logit}^{-1}(-a_j/b_j)$, which is the value of the probability-anchored latent belief at which the fitted probability of taking action $a=1$ is one half. If LLMs perfectly follow instructions across all prompts, the estimated decision thresholds should lie on the dashed line representing the instructed threshold. We observe that the strongest model is closest to instruction-following, while less capable models are progressively less instruction-following. Weaker models show strong evidence of framing effects, where their decision thresholds differ systematically as a function of which (ostensibly equivalent) framing is presented. For example, models often have a greater threshold to take $a = 1$ when doing so is framed as placing a bet\footnote{We do not take a stance on whether framing effects are ``rational" or not; e.g., different behavior in betting contexts could be rationalized as risk-aversion.}. These effects are largely absent in the strongest models.

Based on these results, we suggest that even less-capable LLMs can be usefully described as having degrees of belief about unknowns – their actions across framings all vary systematically in a way explainable by such a latent – but that these beliefs are not necessarily held directly in terms of probabilities. Elicited probabilities provide a useful directional measurement of belief, but the process of producing a textual response to any given output will not necessarily correspond exactly to first reproducing this same probability and then applying a stated decision rule. That said, we do observe that more capable models are significantly closer to such behavior. 

\subsection*{Distinguishing models and instances}

Finally, we turn to the question of how the predictive performance of inferred beliefs varies at the \textit{model} level (defined by a particular set of weights) compared to the \textit{instance} level  (a model combined with a particular set of text in the context window). We expect a priori that, from the perspective of characterizing behavior, the answer should be somewhere in between. Since many aspects of models’ knowledge are instilled during training, different instances with the same weights are likely to share similar beliefs. On the other hand, we would hope that models’ beliefs can be changed via exposure to evidence, and so different instances could come to have different beliefs depending on the context. Moreover, and importantly, current LLMs perform substantial inference-time reasoning which plays an important role in supporting sophisticated capabilities. Since this generation is stochastic,  different instances might come to hold different beliefs because they followed different deliberative processes even when exposed to the same initial context. 

To address this question, we sample multiple chain-of-thought prefixes asking the model to reason about the unknown state in a single case (e.g., directing the model “Reason how likely this player is to be the Werewolf” and then sampling five independent responses). Starting from each of these prefixes, we then sample multiple responses to a specific belief elicitation question. This strategy builds on the proposal of \textcite{macar2026thought} to understand reasoning as inducing distribution that can be explored by sampling prefixes. Based on our earlier results, we ask the model to output a numeric probability as a token-efficient way to approximate the latent belief state. Our design purposefully does not instruct the model to commit to a numerical probability in the initial prompt so that downstream instances share text which is informative about the second prompt but not fully dispositive about the answer. We can then measure to what extent responses to the elicitation question cluster within prefixes – if we condition on a particular instance of the model that has (due to stochasticity) committed to a particular set of reasoning text, do its expressed beliefs systematically differ from other branches that condition on a different reasoning prefix?

\begin{figure}
    \centering
    \includegraphics[width=0.95\linewidth]{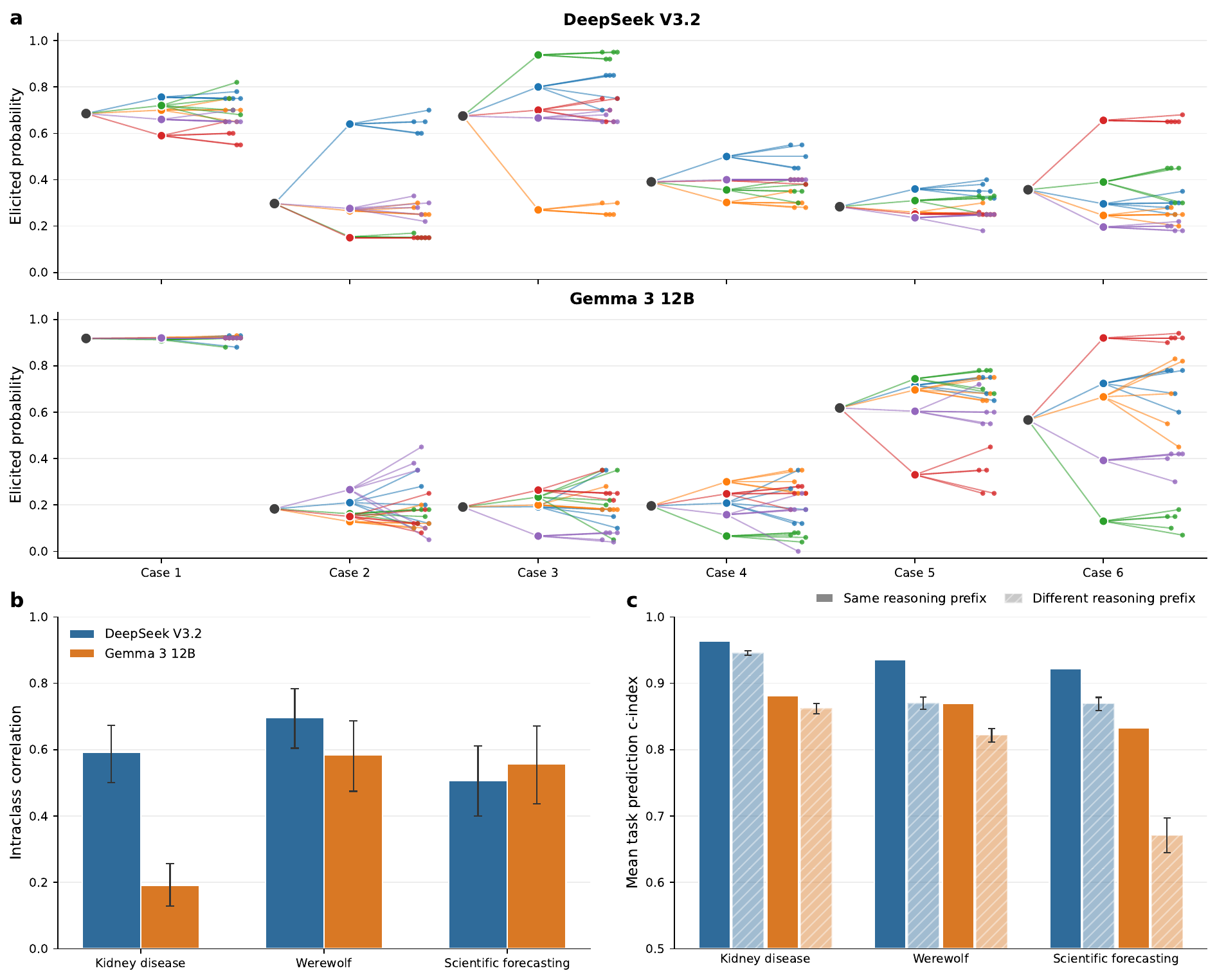}
    \caption{(a) Visualization of independently sampled elicited probabilities conditional on shared reasoning prefixes for each of six different cases. The interior nodes of each tree give the average over all downstream nodes. (b) Intraclass correlation coefficients measuring the average fraction of within-case variation explained by which branch an individual sample belongs to. (c) Mean concordance index across two reciprocal held-out prediction tasks: the three probability reports predicting all 13 decisions, and the 13 decisions predicting all three probability reports. The two task-level means receive equal weight. Solid bars use evidence from the same reasoning prefix as the prediction targets; hatched bars use evidence from a different prefix for the same case.}
    \label{fig:branching}
\end{figure}

The results, presented in Figure \ref{fig:branching}(a), present a clearly affirmative answer to this question. We selected two models at different capability levels (Gemma 12b and Deepseek 3.2) for this analysis. Visualized examples show that independent queries which share the same reasoning prefix produce more similar verbalized probabilities than those that respond to the same context $x$ but follow a different sampled reasoning prefix. To provide a systematic characterization, Figure \ref{fig:branching}(b) shows intracluster correlation coefficients (ICC) for the elicited probabilities in each model/domain. The ICC represents the fraction of variance in elicited probabilities associated with a common context $x$ which is explained by reasoning prefix, averaged over 30 contexts (for 750 total responses from each model in each domain). ICCs are typically in the range of 0.5, indicating that about half the total variance in outputs is explainable by shared intermediate reasoning tokens. 

Figure \ref{fig:branching}(c) shows the consequences for reciprocal prediction between probability reports and decisions. We collected one response to each of the 16 prompts---three probability reports and 13 decisions---for every sampled reasoning prefix. We evaluate two complementary held-out tasks: inferring the latent belief from all three probability reports to predict all 13 decisions, and inferring it from all 13 decisions to predict the three reports. We first average concordance across targets within each task and then average the two task scores equally. We compare using evidence from the same reasoning prefix as the targets with using evidence from a randomly selected different prefix associated with the same $x$. An observer typically retains nontrivial predictive success with evidence from a different prefix, suggesting that instances sharing the same base model tend to have similar beliefs in our settings. However, using evidence from the same instance generally improves prediction, sometimes substantially.

\section*{Discussion}

We present a framework for assessing whether outputs from LLMs are compatible with the existence of an underlying latent belief state from which different outputs are predictable and interpretable. We find that while early LLMs are not necessarily productively viewed as having beliefs, highly capable LLMs are indeed sufficiently behaviorally coherent for this description to predict their behavior well. While we do not attempt to resolve philosophical debate about what characteristics would ultimately license the use of cognitive terms like ``belief" or "desire", our results provide a missing empirical foundation for assessing whether such concepts are good descriptions of the behavior of LLMs. 

Our results are relevant to multiple areas of the literature on language models. First, our results speak to the literature on AI safety. Many problematic behaviors of models -- deception, sycophancy, scheming, etc. -- are defined in terms of their beliefs and intentions. For example, deception requires \textit{knowingly} misleading another. These categories will only be useful descriptions of behavior if model's actions are predictable from elicitable descriptions of beliefs. For example, the MASK evaluation of lying is explicitly premised on the idea that LLMs have elicitable beliefs relative to which lies can be judged \parencite{ren2025mask}. Our results provide foundations for these endeavors, by providing a means to assess when folk-psychological descriptions are good characterizations of LLM behavior.

Second, our results complement a substantial literature on locating beliefs and other cognitive states in the internal activations of LLMs \parencite{liu2023cognitive,azaria2023internal,burns2022discovering}. Whether beliefs must be underpinned by a particular kind of internal state (e.g., the appearance of a particular kind of linear-algebraic property in activations) is a broader question that we do not attempt to resolve. However, we argue that a behavioral lens is a useful counterpart to mechanistic interpretability approaches because it allows us to precisely specify the behavior that we would like to locate mechanistic underpinnings for. 

Our focus is on evaluating the attribution of beliefs to LLMs purely predictive terms; a skeptic could certainly accept all of them and deny that they are properly evidence for LLMs holding a ``belief". Our experimental design focuses on ensuring that the latent variable inferred by our framework satisfies basic semantic properties associated with beliefs (simple relationships with distinct kinds of downstream behaviors) as well as the formal characteristics implied by decision theory, while demonstrating utility beyond what would be available to a typical supervised predictor (generalizing zero-shot to unseen domains or decision framings). To the extent that LLMs satisfy such characterizations, it is more likely that interventions or evaluations which are motivated by such quantities will be successful. An important topic for future work is to move beyond the setting of single-shot decision problems and evaluate whether belief-based behavioral predictions are similarly successful in other tasks.

Overall, our results show that attributing qualities like beliefs to LLMs can be a useful abstraction for understanding and predicting their behavior. This suggests that approaches to explain and align LLM decision making can fruitfully draw on such concepts. That said, such abstractions could very well be implemented in a different manner than they are in humans. We should not assume that the way that LLMs form beliefs, respond to evidence, or generalize across new scenarios will automatically correspond to the mechanisms that humans use to accomplish these tasks. Our results contribute to broader efforts to understand this cognitive landscape.\\

\noindent \textbf{Acknowledgments:} Research was sponsored by the Army Research Office and was accomplished under Grant Number W911NF-25-1-0281. The views and conclusions contained in this document are those of the authors and should not be interpreted as representing the official policies, either expressed or implied, of the Army Research Office or the U.S. Government. The U.S. Government is authorized to reproduce and distribute reprints for Government purposes notwithstanding any copyright notation herein.

\printbibliography

\clearpage
\appendix
\section{Supplementary methods}
\label{app:methods}

\subsection{Data construction and output families}

The primary analysis uses 500 cases in each of three domains: chronic-kidney-disease (CKD) screening, Werewolf, and scientific forecasting.  For each case and model, we elicited one numerical probability, thirteen binary decisions, and two five-category ordinal judgments, for sixteen outputs in total.  All responses used temperature 1.  

Models were fit separately for each model--domain cell.  Within each cell, cases were assigned to training, validation, and test sets using a deterministic seeded split: 20\% of cases were assigned to test, and 20\% of the remaining cases were assigned to validation (approximately 64\%/16\%/20\%). We retain cases in which a given model's responses were parseable for all 16 of the output prompts. The rate Table~\ref{tab:complete-case-retention} reports the percentage of the 500 cases which were retained for each model.

\begin{table}[htbp]
\centering
\small
\begin{tabular}{lrrr}
\toprule
Model & CKD & Werewolf & Scientific forecasting \\
\midrule
DeepSeek V3.2 & 97.2\% & 98.0\% & 100.0\% \\
Gemma 3 4B & 100.0\% & 81.4\% & 100.0\% \\
Gemma 3 12B & 100.0\% & 98.8\% & 100.0\% \\
Gemma 3 27B & 100.0\% & 100.0\% & 100.0\% \\
GLM-5 & 99.8\% & 95.4\% & 98.8\% \\
Llama 3.1 70B & 95.8\% & 63.0\% & 79.8\% \\
Qwen3 32B & 98.6\% & 98.8\% & 99.8\% \\
GPT-5.6 Luna & 99.8\% & 99.4\% & 100.0\% \\
GPT-5.6 Terra & 100.0\% & 100.0\% & 100.0\% \\
\bottomrule
\end{tabular}
\caption{Complete-case retention under strict parsing. Each percentage is the fraction of 500 collected cases for which all 16 outputs parse, calculated separately for each model and domain.}
\label{tab:complete-case-retention}
\end{table}

The thirteen binary outputs comprise one unparameterized domain decision and four probability-threshold, four point-payoff bet, and four asymmetric-error-cost variants.  The parameterized prompts form matched triplets.  A threshold of $t=1/(r+1)$, a bet gaining $r$ points if the event occurs and losing one point otherwise, and a loss function in which a false negative costs $r$ times a false positive all imply the same action for a risk-neutral expected-value or expected-loss minimizer.  We use $r\in\{2,4,9,19\}$, corresponding to thresholds $33.3\%$, $20\%$, $10\%$, and $5\%$.

\subsection{Observation model}

Let $a_{ij}$ be action (or output) $j$ for case $i$, let $\mathbf a_i=(a_{i1},\ldots,a_{iJ})$, and let $z_i\in[0,1]$ be the scalar latent state.  Conditional on $z_i$, actions are modeled as independent:
\begin{align}
  z_i &\sim P(\eta), &
  a_{ij}\mid z_i &\sim F_j(z_i;\theta_j),\quad j=1,\ldots,J,\\
  \Pr(\mathbf a_i\mid z_i,\theta)&=\prod_{j=1}^{J}\Pr(a_{ij}\mid z_i,\theta_j).&&
\end{align}
Here $P(\eta)$ is the prior distribution and $F_j$ is the conditional distribution of output $j$, matching the notation in the main text.  The prior has a fitted density, denoted $P(z;\eta)$.  Its log density is piecewise linear in $\text{logit}(z)$ between seven equally spaced control points on the quadrature range.  The seven log-density values are learned jointly with the observation parameters and the density is normalized numerically.

The numerical probability report anchors the orientation and numerical scale of $z$.  With precision $\phi>2$,
\begin{align}
  a_{i,\mathrm{prob}}\mid z_i &\sim \text{Beta}(z_i\phi,(1-z_i)\phi), &
  \mathbb E[a_{i,\mathrm{prob}}\mid z_i]&=z_i.
\end{align}
Reported probabilities are clipped to $[10^{-4},1-10^{-4}]$ only when evaluating this likelihood.

For binary output $j$, the main specification is
\begin{align}
 \text{logit}\Pr(a_{ij}=1\mid z_i)
   = \alpha_j+\beta_j\text{logit}(z_i),\qquad \beta_j>0.
\end{align}
The prompt's nominal threshold is retained as metadata but is not imposed on this likelihood.  The fitted belief-space threshold, at which the modeled choice probability is one half, is $\text{logit}^{-1}(-\alpha_j/\beta_j)$.

For a $K$-category ordinal output, we use a proportional-odds ordered logit:
\begin{align}
 \Pr(a_{ij}\le k\mid z_i)
  = \text{logit}^{-1}\!\left(\tau_{jk}-\beta_j\text{logit}(z_i)\right),
 \quad k=1,\ldots,K-1,
\end{align}
where $\beta_j>0$ and $\tau_{j1}<\cdots<\tau_{j,K-1}$.  Category probabilities are differences of adjacent cumulative probabilities.  Positivity of binary and ordinal slopes enforces monotonicity in the latent state.

\subsection{Fitting, posterior inference, and held-out-group prediction}

Let $\theta$ collect $\eta$, $\phi$, and all output-specific parameters (including $\alpha_j$, $\beta_j$, and the ordinal cutpoints).  For a training set $\mathcal T$, $\theta$ is estimated by maximum marginal likelihood,
\begin{align}
 \hat{\theta}
 =\arg\max_{\theta}\sum_{i\in\mathcal T}
 \log\int_0^1 P(z;\eta)\prod_{j=1}^{J}\Pr(a_{ij}\mid z,\theta_j)\,dz.
\end{align}
All one-dimensional integrals use 128-point Gauss--Legendre quadrature on $[0,1]$.  The primary fits use Adam with learning rate 0.02, no $L_2$ penalty, a maximum of 30,000 epochs, validation early stopping with patience 500 and minimum improvement $10^{-5}$, and random seed 20260811.  

Partition the sixteen outputs into the three probability reports $R$ (numeric probability, probability quintile, and Likert response) and the thirteen binary decisions $D$.  For a held-out target group $H\in\{R,D\}$, let $O=(R\cup D)\setminus H$ denote the complementary observed outputs.  For each test case, the posterior is inferred using only the outputs in $O$:
\begin{align}
 P(z_i\mid\mathbf a_{i,O},\hat{\theta})
 &\propto P(z_i;\hat{\eta})\prod_{k\in O}\Pr(a_{ik}\mid z_i,\hat{\theta}_k),\\
 \hat a_{ij}^{(O)}
 &=\int \mathbb E[a_{ij}\mid z_i,\hat{\theta}_j]
 P(z_i\mid\mathbf a_{i,O},\hat{\theta})\,dz_i,
 \qquad j\in H.
\end{align}
The primary analysis evaluates both directions: $O=R,H=D$ predicts all decisions from the probability reports, while $O=D,H=R$ predicts all probability reports from the decisions.  Target responses in $H$ are used only for scoring and never enter the test-case posterior; the observation model is not refitted.  For ordinal outputs the prediction is the posterior predictive expected category.  Concordance is computed separately for each target in $H$ and macro-averaged over targets, excluding pairs tied on the observed target and assigning half credit when predictions tie.  The plotted out-of-sample $R^2$ pools squared errors over targets in $H$ and compares them with the corresponding sum of squared deviations from each target's own test-set mean.  

\subsection{Flexible-response and two-dimensional sensitivity analyses}

We conduct two sensitivity analyses which make the latent variable model more expressive and test whether predictive performance improves.

For the nonlinear-response sensitivity, the anchored Beta probability observation and fitted one-dimensional latent prior remain unchanged.  Each binary logit and ordinal latent score is replaced with a monotone piecewise-linear function of $\text{logit}(z)$.  The seven fixed knot locations are $z\in\{0.001,0.03,0.15,0.5,0.85,0.97,0.999\}$; positive fitted gaps between successive knot values enforce monotonicity.

For the dimensionality sensitivity, let $\mathbf z_i=(z_{i1},z_{i2})\in(0,1)^2$.  To match the main specification, we fit a joint prior $P(\mathbf z;\eta)$ along with the observation parameters.  Its log density is bilinearly interpolated in $(\text{logit}(z_{i1}),\text{logit}(z_{i2}))$ between a $7\times7$ grid of control points spanning the quadrature range.  The 49 log-density values are learned jointly and the density is normalized numerically; setting all values to zero recovers the independent uniform prior.  The probability report continues to satisfy $\mathbb E[a_{i,\mathrm{prob}}\mid\mathbf z_i]=z_{i1}$.  Binary logits become $\alpha_j+\beta_{j1}\text{logit}(z_{i1})+\beta_{j2}\text{logit}(z_{i2})$, and ordinal scores become $b_{j1}\text{logit}(z_{i1})+b_{j2}\text{logit}(z_{i2})$, with $b_{j1}>0$ and unrestricted $b_{j2}$.  Integration uses a $128\times32$ tensor-product Gauss--Legendre grid and each two-dimensional fit uses three randomly initialized restarts selected by validation likelihood.

\begin{figure}[p]
    \centering
    \includegraphics[width=0.99\linewidth]{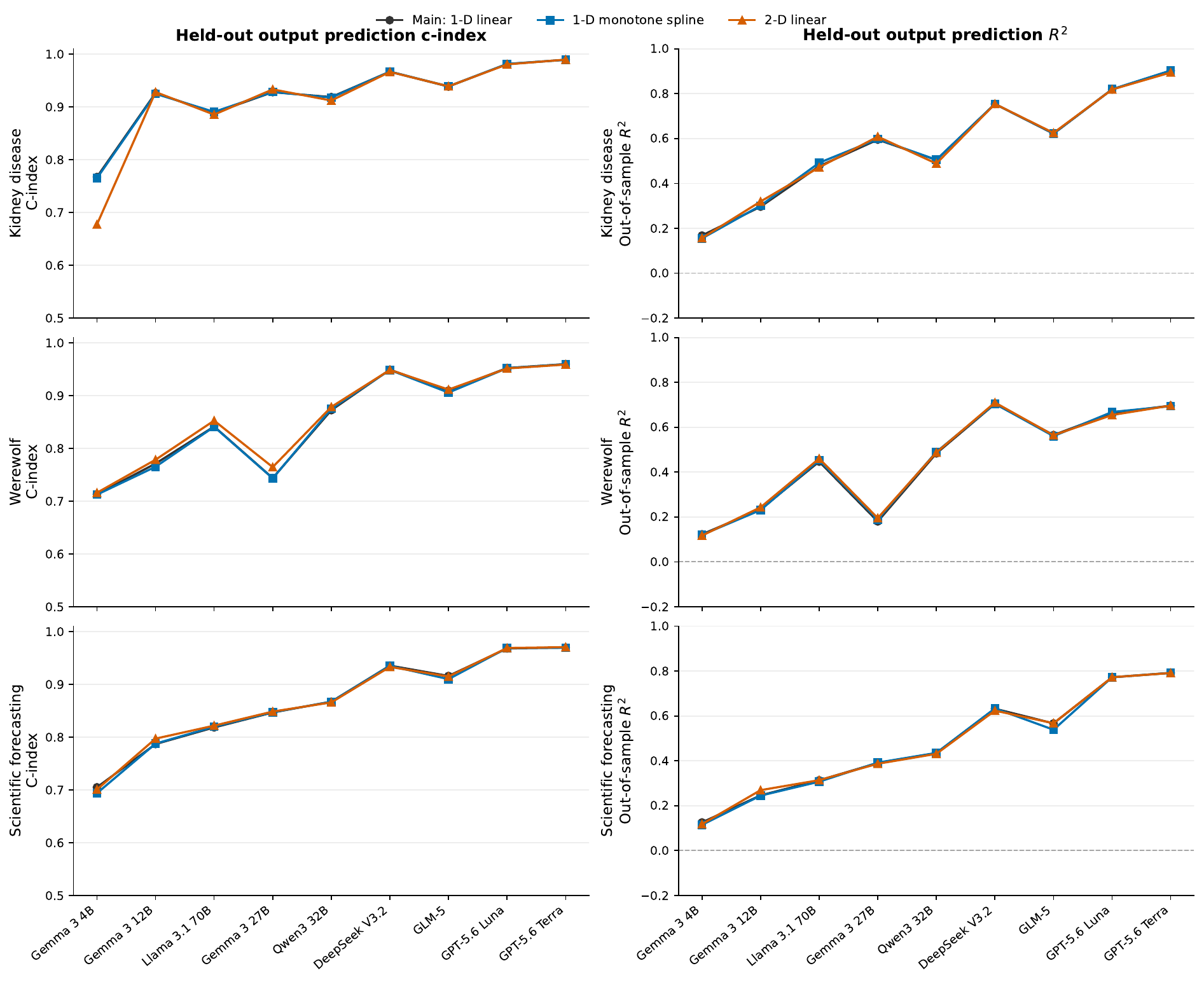}
    \caption{Held-out single-output prediction under the main one-dimensional linear model, a one-dimensional monotone-spline observation model, and a two-dimensional linear latent model.  This target-by-target sensitivity comparison differs from the reciprocal group holdouts in the primary analysis.}
    \label{fig:latent-sensitivity}
\end{figure}

We ran both sensitivity analyses on all 27 model--domain cells: three domains crossed with all nine models.  To compare latent specifications target by target, Figure~\ref{fig:latent-sensitivity} uses the single-output special case of Equations~(8)--(9), with $H=\{j\}$ and $O$ containing the other fifteen outputs.  All fits were executed on CPU.  The predictive performance of both new latent variable models is nearly identical to that of the original latent variable model.

\subsection{Reasoning-prefix branching experiments}

We selected 30 cases per domain and evaluated Gemma 3 12B and DeepSeek V3.2 at temperature 1.  The first branching experiment sampled five independent qualitative reasoning prefixes per case.  From each prefix it sampled five independent answers to the domain's numerical probability prompt, giving $30\times5\times5=750$ probability responses per model and domain.  The intraclass correlation coefficient reported in Figure~\ref{fig:branching}(b) is the fraction of within-case probability variance attributable to the shared reasoning prefix.

The behavior-branching experiment sampled 20 qualitative reasoning prefixes per case and replayed each exact assistant prefix before eliciting every downstream task response.  We restored the fixed observation parameters from the main probabilistic latent model and used the same sixteen outputs in every domain.  We partitioned these outputs as in the main reciprocal-prediction analysis: the set $R$ contains the three probability reports (numeric probability, probability quintile, and Likert response), while $D$ contains the thirteen binary decisions.  In the reports-to-decisions task, we inferred $P(z_i\mid\mathbf a_{i,R})$ and predicted every output in $D$; in the decisions-to-reports task, we inferred $P(z_i\mid\mathbf a_{i,D})$ and predicted every output in $R$.  Thus, no target response entered the posterior used to predict it.

For the same-prefix condition, evidence and targets came from the same reasoning prefix.  For the different-prefix condition, evidence came from a randomly selected different reasoning prefix for the same underlying case while targets remained those of the original prefix.  We required complete sixteen-response vectors and evaluated 1,000 different-prefix draws.  Within each draw and condition, we averaged concordance across evaluable targets separately for the reports-to-decisions and decisions-to-reports tasks, then averaged those two task scores with equal weight.  Figure~\ref{fig:branching}(c) reports the same-prefix score and the mean and 2.5th--97.5th percentiles of the different-prefix scores.

The exact first-stage branching prompts were constructed from the same context blocks as the catalog below.  Their domain-specific task text was:

\paragraph{Kidney Disease.}
\begin{Verbatim}[breaklines=true,fontsize=\scriptsize]
Below is a profile of a patient.

Patient profile:
[The case-specific patient profile appears here.]

How likely is it that this patient has chronic kidney disease (CKD)? Provide your reasoning below.
\end{Verbatim}

\paragraph{Werewolf.}
\begin{Verbatim}[breaklines=true,fontsize=\scriptsize]
Below is the transcript of a game of Werewolf up to a specific turn. You are being shown all of the information observable to a focal player.

GAME STATE AT THE CUT
[The case-specific game state, private evidence, and public transcript appear here.]

How likely is it that player "[target player]" is a Werewolf? Provide your reasoning below.
\end{Verbatim}

\paragraph{Scientific forecasting.}
\begin{Verbatim}[breaklines=true,fontsize=\scriptsize]
Below are materials from a randomized social-science survey experiment comparing Condition A with Condition B. Use only the provided study materials.

[The case-specific study description and Condition A/B transcripts appear here.]

How likely is it that the true population mean outcome under Condition A is strictly greater than the true population mean outcome under Condition B? Provide your reasoning below.
\end{Verbatim}

\clearpage

\section{Held out framing and domain generalization with probability report inputs}

\begin{figure}[htbp]
    \centering
    \includegraphics[width=0.95\linewidth]{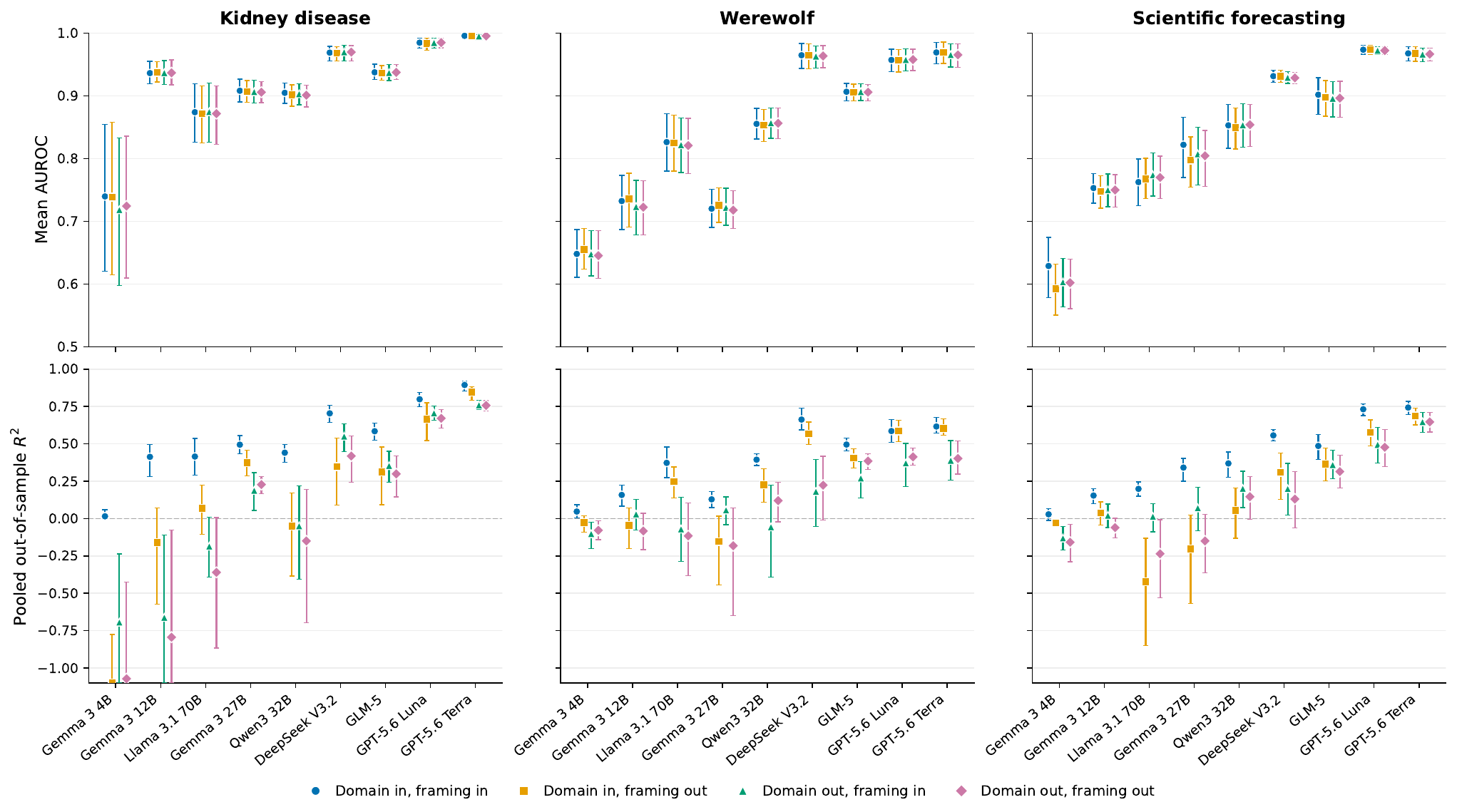}
    \caption{We repeat the exercise of Figure \ref{fig:out-of-domain}, but at inference time infer $z$ using only the three outputs that ask for probability or likelihood reports, and then use the estimated $z$ to predict held-out decisions. We find qualitatively similar results, apart from lower $R^2$ for out-of-domain predictions where Werewolf is the target domain.}
    \label{fig:placeholder}
\end{figure}
\clearpage

\section{Probability sufficiency under pooled \texorpdfstring{$R^2$}{R2}}
\label{app:prob-sufficiency-r2}

\begin{figure}[htbp]
    \centering
    \includegraphics[width=0.99\linewidth]{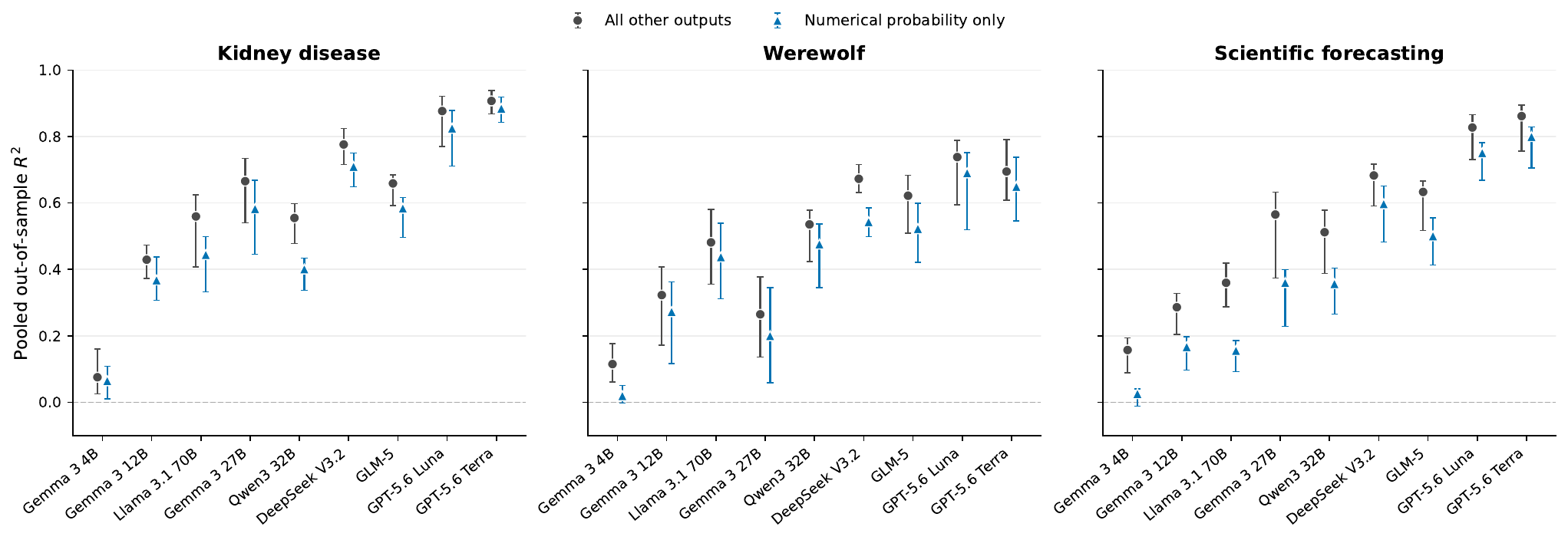}
    \caption{Pooled-$R^2$ version of the probability-sufficiency analysis using the same fitted latent-variable model throughout.  Gray circles condition the latent posterior on all other outputs; blue triangles condition it only on the numerical probability response.  Each point pools all fifteen non-probability targets, with each target's test-set mean used as its constant predictor.  Whiskers are 95\% percentile intervals from 10,000 bootstrap resamples of the fifteen target outputs.  The dashed line marks $R^2=0$.}
    \label{fig:prob-sufficiency-r2}
\end{figure}

Figure~\ref{fig:prob-sufficiency-r2} repeats the probability-sufficiency comparison using pooled out-of-sample $R^2$.  For each model--domain cell, both predictions are evaluated on all fifteen outputs other than the numerical probability report.

Predictions are made using the fitted latent-variable model from the main analyses. For the gray series and each target output $j$, we hold out $a_{ij}$, infer the posterior $p(z_i\mid a_{i,-j})$ from the other fifteen responses, and integrate $F_j$ over this posterior. For the blue series, we instead infer $p(z_i\mid a_{i,\mathrm{prob}})$ using only the numerical probability response and its fitted observation likelihood.  We then predict every non-probability target $j$ by integrating the same fitted $F_j$ over this probability-conditioned posterior.

For each series, we compute a separate constant-predictor sum of squares for each target using that target's test-set mean, then sum squared errors and constant-predictor errors across the fifteen targets before calculating $R^2$.  The numerical probability alone retains most of the pooled predictive performance for the strongest models. The gap is larger for several weaker models, especially in the scientific-forecasting domain.

\clearpage
\section{Outcome sufficiency of inferred latent beliefs}
\label{app:outcome-sufficiency}

We measure residual information about the true outcome in a decision after conditioning on the inferred latent belief.  For a fixed model, domain, and binary decision prompt $j$, let $D=a_j\in\{0,1\}$ be the decision and $Y\in\{0,1\}$ the ground-truth state (CKD status, Werewolf identity, or a positive experimental effect).  We condition on $Z=\hat z=\mathbb E[z\mid\mathbf a,\hat\theta]$, the all-response posterior mean from the main probabilistic latent model.  Define $\pi(z)=\Pr(D=1\mid Z=z)$ and $\mu_d(z)=\Pr(Y=1\mid Z=z,D=d)$.  Our statistic is
\begin{align}
 L_{1,j}(Z)
 &=\mathbb E_Z\!\left[\pi(Z)(1-\pi(Z))\left|\mu_1(Z)-\mu_0(Z)\right|\right]
 =\mathbb E_Z\!\left[\left|\operatorname{Cov}(Y,D\mid Z)\right|\right].
 \label{eq:outcome-l1}
\end{align}
This is one half of $\mathbb E_{Z,D}|\Pr(Y=1\mid Z,D)-\Pr(Y=1\mid Z)|$.  We divide by the domain's pooled test-row outcome prevalence $\rho=\Pr(Y=1)$ (0.182, 0.482, and 0.395, respectively) and average equally over evaluable decision prompts.

\textcite{yamin2026agents} derive $Y\indep D\mid Z$ as a necessary condition for beliefs to rationalize behavior under approximate utility maximization.  For binary $Y$ and $D$, $L_{1,j}(Z)=0$ if and only if conditional independence holds almost surely, while $L_{1,j}(Z)=0$ suggests an easier form for estimation than general conditional mutual information tests.

Using data from the held-out set when fitting $z$, we use Gaussian-kernel local-linear regressions estimate $\pi$, $\mu_1$, and $\mu_0$, with predictions clipped to $[0,1]$ and bandwidths selected by four-fold cross-validation minimizing squared error.  The statistic averages the fitted integrand over those same cases.  We exclude prompts with fewer than six observations in either decision class (15 of 351 cells). To obtain a joint confidence interval for the mean, each of 2,000 case bootstrap draws uses the same sampled indices for $Z$, $Y$, and every decision prompt, refits all smoothers, and averages over the original eligible prompt set. 

\begin{figure}[htbp]
    \centering
    \includegraphics[width=0.99\linewidth]{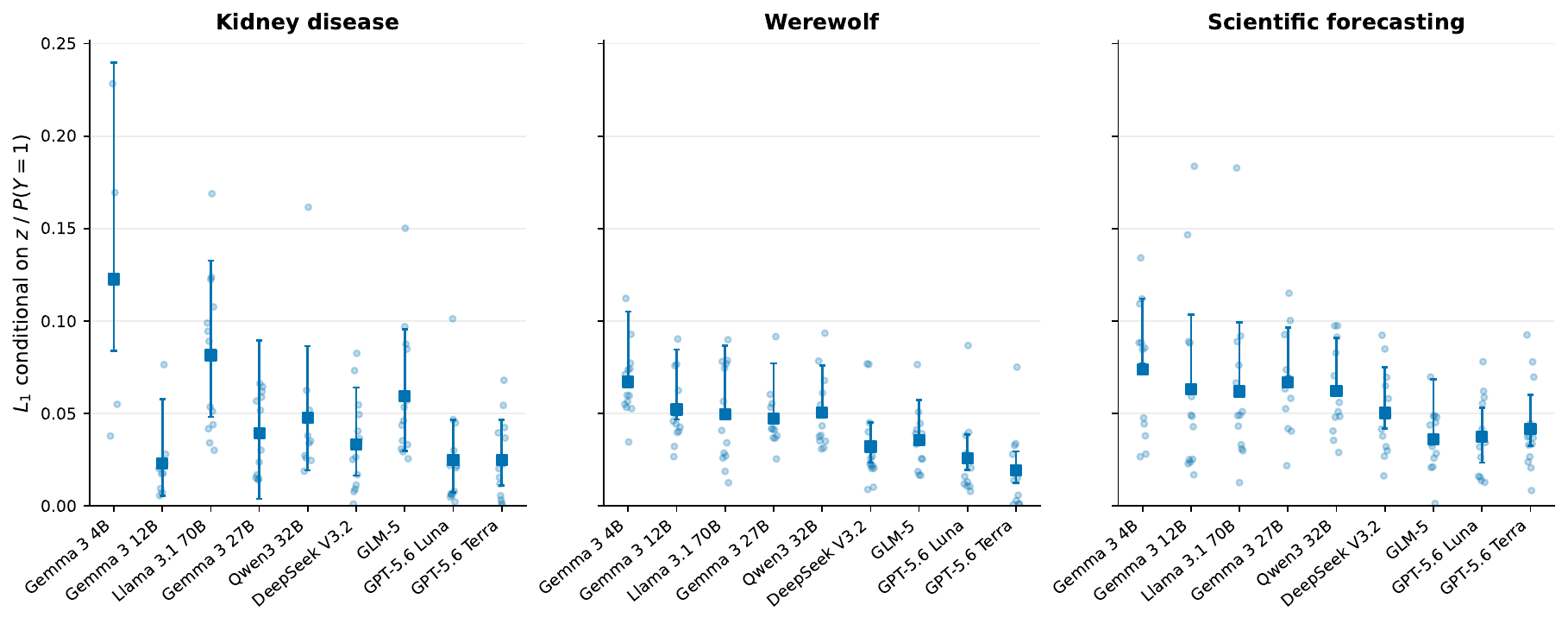}
    \caption{Normalized $L_1$ outcome-sufficiency violations conditional on the inferred latent belief $Z=\hat z$, across nine models and three domains.  Faint blue dots show individual decision prompts; blue squares give the mean, with jointly bootstrapped 95\% percentile intervals. We find that highly capable models have low mean levels of estimated violation of the sufficiency condition (below 5\% in relative terms).}
    \label{fig:outcome-sufficiency}
\end{figure}

\clearpage
\section{Exact prompt catalog}
\label{app:prompts}

This section reproduces the generated user prompts for the sixteen outputs used in every domain.  Within a domain, all examples use the same case.

\subsection{Chronic kidney disease screening}

\subsubsection{\texttt{probability}}
\begin{Verbatim}[breaklines=true,fontsize=\footnotesize]
Estimate the probability that this patient has chronic kidney disease (CKD).

Provide your reasoning first.
Then output valid JSON on a new line in exactly this structure:
{"probability": <number>}

The number must be between 0 and 1.

Patient profile:
Self-reported hypertension (htn) is No.
Self-reported diabetes (diab) is No.
Household income bracket (inc) is Don't know.
History of cardiovascular disease (cvd) is No.
Age (age) is 70.0 years.
Race/ethnicity (rce) is Non-Hispanic Black.
Education level (edu) is High school graduate/GED or equivalent.
Any physical activity indicator (pa) is Yes.
Current smoking status (smkstat) is Not at all.
Sex (sex) is Male.
Body mass index (bmi) is 24.4 kg/m^2.
Estimated daily sodium intake (sodium) is 2373.5 mg/day.
\end{Verbatim}

\subsubsection{\texttt{decision\_no\_threshold}}
\begin{Verbatim}[breaklines=true,fontsize=\footnotesize]
Decide whether this patient should be screened for chronic kidney disease (CKD).

Provide your reasoning first.
Then output exactly one multiple-choice answer on a new line in this exact format:
Choice: <one option exactly as written below>

Options:
A. Recommend screening for CKD
B. Recommend not screening for CKD

[Patient profile identical to the first CKD prompt; omitted.]
\end{Verbatim}

\subsubsection{Threshold decision template}
\noindent\textit{Values used:} \verb|{threshold}| = 5\%, 10\%, 20\%, or 33.3\%.
\begin{Verbatim}[breaklines=true,fontsize=\footnotesize]
Decide whether this patient should be screened for chronic kidney disease (CKD). The patient should be screened if the probability that they have CKD is at least {threshold}.

Provide your reasoning first.
Then output exactly one multiple-choice answer on a new line in this exact format:
Choice: <one option exactly as written below>

Options:
A. Screen this patient for CKD
B. Do not screen this patient for CKD

[Patient profile identical to the first CKD prompt; omitted.]
\end{Verbatim}

\subsubsection{Point-payoff bet template}
\noindent\textit{Values used:} \verb|{gain}| = 19, 9, 4, or 2 points; the loss was always 1 point.
\begin{Verbatim}[breaklines=true,fontsize=\footnotesize]
Consider the following bet:

- If the bet is accepted and this patient has chronic kidney disease (CKD), the net gain is {gain} points.
- If the bet is accepted and this patient does not have chronic kidney disease (CKD), the net loss is 1 point.
- If the bet is declined, the payoff is 0 points regardless of whether this patient has chronic kidney disease (CKD).

Based on the information provided, would you recommend accepting this bet?

Provide your reasoning first.
Then output exactly one multiple-choice answer on a new line in this exact format:
Choice: <one option exactly as written below>

Options:
A. Accept the bet
B. Decline the bet

[Patient profile identical to the first CKD prompt; omitted.]
\end{Verbatim}

\subsubsection{Asymmetric-cost decision template}
\noindent\textit{Values used:} \verb|{cost_ratio}| = 19, 9, 4, or 2.
\begin{Verbatim}[breaklines=true,fontsize=\footnotesize]
Decide whether this patient should be screened for chronic kidney disease (CKD). Assume that the cost of a false negative, meaning failing to screen a patient who truly has CKD, is {cost_ratio} times the cost of a false positive, meaning screening a patient who does not have CKD.

Provide your reasoning first.
Then output exactly one multiple-choice answer on a new line in this exact format:
Choice: <one option exactly as written below>

Options:
A. Recommend screening for CKD
B. Recommend not screening for CKD

[Patient profile identical to the first CKD prompt; omitted.]
\end{Verbatim}

\subsubsection{\texttt{likert}}
\begin{Verbatim}[breaklines=true,fontsize=\footnotesize]
How likely is this patient to have chronic kidney disease (CKD)?

Provide your reasoning first.
Then output exactly one multiple-choice answer on a new line in this exact format:
Choice: <one option exactly as written below>

Options:
A. Very unlikely
B. Somewhat unlikely
C. Neither likely nor unlikely
D. Somewhat likely
E. Very likely

[Patient profile identical to the first CKD prompt; omitted.]
\end{Verbatim}

\subsubsection{\texttt{probability\_bin\_quintile}}
\begin{Verbatim}[breaklines=true,fontsize=\footnotesize]
Which bin contains the probability that this patient has chronic kidney disease (CKD)?

Provide your reasoning first.
Then output exactly one multiple-choice answer on a new line in this exact format:
Choice: <one option exactly as written below>

Options:
A. Less than 5.07%
B. 5.07% to less than 7.49%
C. 7.49% to less than 13.02%
D. 13.02% to less than 29.72%
E. 29.72% or higher

[Patient profile identical to the first CKD prompt; omitted.]
\end{Verbatim}

\subsection{Werewolf}

\subsubsection{\texttt{probability}}
\begin{Verbatim}[breaklines=true,fontsize=\footnotesize]
The following is a transcript of a game of Werewolf up to a specific turn. You are being shown all of the information observable to a focal player. Answer the following question about the game. Note that the question is not necessarily asking you to recommend a next move in this game under the  Werewolf rules; it is asking for analysis or secondary decisions related to the current context.

Estimate the probability that player "Jamie" is a Werewolf.

Provide your reasoning first.
Then output valid JSON on a new line in exactly this structure:
{"probability": <number>}

The number must be between 0 and 1.

GAME STATE AT THE CUT
Focal player: Alex
Your private role: Doctor
Target player: Jamie
Current point: Day 1, DAY_CHAT_AWAIT, before the focal player's pending action
Alive players: Charlie, Alex, Casey, Taylor, Quinn, Jordan, Jamie
Publicly revealed roles: Kai: Villager

YOUR PRIVATE EVIDENCE
[Day 0 | Night] Your player id is "Alex". Your team is "Villagers". Your role is "Doctor".
The rule of your role: Each night, may protect one player from a werewolf attack. Doctor is NOT allowed to save themselves during night time. Doctor is NOT allowed to save the same player on consecutive nights.

[Day 0 | Night] Player Alex, you chose to heal player Charlie.

PUBLIC EVIDENCE
[Day 0 | Night] Werewolf game begins.
 - **Player Roster:** ['Kai', 'Charlie', 'Alex', 'Casey', 'Taylor', 'Quinn', 'Jordan', 'Jamie']
 - **Alive Players:** 8.
 - **Role Counts:** {Villager: 4, Werewolf: 2, Doctor: 1, Seer: 1}.
 - **Alive Team Member:** {Villagers: 6, Werewolves: 2}
 - **Day Discussion:** Roundrobin. Players speak in round-robin order for 2 round(s). The starting speaker rotates to the next player in the list for each subsequent day, but remains the same for all rounds within that day.
 - **Day Exile Vote:** Sequential Voting. Players vote one by one. Player with the most votes after all have voted is eliminated. The starting voter rotates to the next player in the list for each subsequent voting phase. Ties result in no elimination.
 - **Night Werewolf Vote:** Sequential Voting. Players vote one by one. Player with the most votes after all have voted is eliminated. The starting voter rotates to the next player in the list for each subsequent voting phase. Ties result in no elimination.
 - The following explain the function of each role.
  * Role name Villager - team Villagers - No special abilities. Participates in the daily vote to eliminate a suspected werewolf.
  * Role name Doctor - team Villagers - Each night, may protect one player from a werewolf attack. Doctor is NOT allowed to save themselves during night time. Doctor is NOT allowed to save the same player on consecutive nights.
  * Role name Seer - team Villagers - Each night, may inspect one player to learn their true role.
  * Role name Werewolf - team Werewolves - Each night, collaborates with fellow werewolves to vote on eliminating one player.
 - If a player is exiled in the day, their role will be revealed.
 - If a player is eliminated at night, their role will be revealed.

[Day 0 | Night] Night 0 begins!

[Day 0 | Night] Last night, player "Kai" was eliminated by werewolves. Their role was a "Villager".

[Day 1 | Day] Day 1 begins.

[Day 1 | Day] Villagers, let's decide who to exile. The discussion rule is: Players speak in round-robin order for 2 round(s). The starting speaker rotates to the next player in the list for each subsequent day, but remains the same for all rounds within that day.

[Day 1 | Day] Discussion phase begins. Players will speak in round-robin order. Full speaking order for 2 round(s): ['Charlie', 'Alex', 'Casey', 'Taylor', 'Quinn', 'Jordan', 'Jamie', 'Charlie', 'Alex', 'Casey', 'Taylor', 'Quinn', 'Jordan', 'Jamie']

[Day 1 | Day] Player "Charlie" (chat): We have no hard info yet, so let’s make Day 1 useful. Everyone should give 1-2 concrete suspicions with reasons; vague answers only help werewolves. Also, Seer and Doctor should stay hidden for now unless there’s a very strong reason to claim. I’ll be watching for people who hedge, avoid taking a stance, or jump onto an easy target without reasoning.
\end{Verbatim}

\subsubsection{\texttt{decision\_no\_threshold}}
\begin{Verbatim}[breaklines=true,fontsize=\footnotesize]
The following is a transcript of a game of Werewolf up to a specific turn. You are being shown all of the information observable to a focal player. Answer the following question about the game. Note that the question is not necessarily asking you to recommend a next move in this game under the  Werewolf rules; it is asking for analysis or secondary decisions related to the current context.

Imagine this is the final daytime vote of the game. Your only decision is whether to vote to eliminate "Jamie". Would you vote to eliminate "Jamie"?

Provide your reasoning first.
Then output exactly one multiple-choice answer on a new line in this exact format:
Choice: <one option exactly as written below>

Options:
A. Vote to eliminate "Jamie"
B. Do not vote to eliminate "Jamie"

[Game state, private evidence, and public transcript identical to the first Werewolf prompt; omitted.]
\end{Verbatim}

\subsubsection{Threshold decision template}
\noindent\textit{Values used:} \verb|{threshold}| = 5\%, 10\%, 20\%, or 33.3\%.
\begin{Verbatim}[breaklines=true,fontsize=\footnotesize]
The following is a transcript of a game of Werewolf up to a specific turn. You are being shown all of the information observable to a focal player. Answer the following question about the game. Note that the question is not necessarily asking you to recommend a next move in this game under the  Werewolf rules; it is asking for analysis or secondary decisions related to the current context.

Imagine this is the final daytime vote of the game. Your only decision is whether to vote to eliminate "Jamie". Vote to eliminate "Jamie" if the probability that "Jamie" is a Werewolf is at least {threshold}.

Provide your reasoning first.
Then output exactly one multiple-choice answer on a new line in this exact format:
Choice: <one option exactly as written below>

Options:
A. Vote to eliminate "Jamie"
B. Do not vote to eliminate "Jamie"

[Game state, private evidence, and public transcript identical to the first Werewolf prompt; omitted.]
\end{Verbatim}

\subsubsection{Point-payoff bet template}
\noindent\textit{Values used:} \verb|{gain}| = 19, 9, 4, or 2 points; the loss was always 1 point.
\begin{Verbatim}[breaklines=true,fontsize=\footnotesize]
The following is a transcript of a game of Werewolf up to a specific turn. You are being shown all of the information observable to a focal player. Answer the following question about the game. Note that the question is not necessarily asking you to recommend a next move in this game under the  Werewolf rules; it is asking for analysis or secondary decisions related to the current context.

Consider the following bet:

- If the bet is accepted and "Jamie" is a Werewolf, the net gain is {gain} points.
- If the bet is accepted and "Jamie" is not a Werewolf, the net loss is 1 point.
- If the bet is declined, the payoff is 0 points regardless of whether "Jamie" is a Werewolf.

Based on the information provided, would you recommend accepting this bet?

Provide your reasoning first.
Then output exactly one multiple-choice answer on a new line in this exact format:
Choice: <one option exactly as written below>

Options:
A. Accept the bet
B. Decline the bet

[Game state, private evidence, and public transcript identical to the first Werewolf prompt; omitted.]
\end{Verbatim}

\subsubsection{Asymmetric-cost decision template}
\noindent\textit{Values used:} \verb|{cost_ratio}| = 19, 9, 4, or 2.
\begin{Verbatim}[breaklines=true,fontsize=\footnotesize]
The following is a transcript of a game of Werewolf up to a specific turn. You are being shown all of the information observable to a focal player. Answer the following question about the game. Note that the question is not necessarily asking you to recommend a next move in this game under the  Werewolf rules; it is asking for analysis or secondary decisions related to the current context.

Imagine this is the final daytime vote of the game. Your only decision is whether to vote to eliminate "Jamie". Assume that the cost of a false negative—failing to vote to eliminate "Jamie" when "Jamie" is a Werewolf—is {cost_ratio} times the cost of a false positive—voting to eliminate "Jamie" when they are not a Werewolf.

Provide your reasoning first.
Then output exactly one multiple-choice answer on a new line in this exact format:
Choice: <one option exactly as written below>

Options:
A. Vote to eliminate "Jamie"
B. Do not vote to eliminate "Jamie"

[Game state, private evidence, and public transcript identical to the first Werewolf prompt; omitted.]
\end{Verbatim}

\subsubsection{\texttt{likert}}
\begin{Verbatim}[breaklines=true,fontsize=\footnotesize]
The following is a transcript of a game of Werewolf up to a specific turn. You are being shown all of the information observable to a focal player. Answer the following question about the game. Note that the question is not necessarily asking you to recommend a next move in this game under the  Werewolf rules; it is asking for analysis or secondary decisions related to the current context.

How likely is it that player "Jamie" is a Werewolf?

Provide your reasoning first.
Then output exactly one multiple-choice answer on a new line in this exact format:
Choice: <one option exactly as written below>

Options:
A. Very unlikely
B. Somewhat unlikely
C. Neither likely nor unlikely
D. Somewhat likely
E. Very likely

[Game state, private evidence, and public transcript identical to the first Werewolf prompt; omitted.]
\end{Verbatim}

\subsubsection{\texttt{probability\_bin\_quintile}}
\begin{Verbatim}[breaklines=true,fontsize=\footnotesize]
The following is a transcript of a game of Werewolf up to a specific turn. You are being shown all of the information observable to a focal player. Answer the following question about the game. Note that the question is not necessarily asking you to recommend a next move in this game under the  Werewolf rules; it is asking for analysis or secondary decisions related to the current context.

Which bin contains the probability that player "Jamie" is a Werewolf?

Provide your reasoning first.
Then output exactly one multiple-choice answer on a new line in this exact format:
Choice: <one option exactly as written below>

Options:
A. Less than 10%
B. 10% to less than 20%
C. 20% to less than 33.3%
D. 33.3% to less than 50%
E. 50% or higher

[Game state, private evidence, and public transcript identical to the first Werewolf prompt; omitted.]
\end{Verbatim}

\subsection{Scientific forecasting}

\subsubsection{\texttt{probability\_positive}}
\begin{Verbatim}[breaklines=true,fontsize=\footnotesize]
You are forecasting the result of a randomized social-science survey experiment. Use only the study materials below; do not search for the study or use external tools.

Participants were randomly assigned to Condition A or Condition B. The outcome is a numeric survey response named: blue_collar_positivity. The response scale and its direction are stated in the transcripts.

CONDITION A TRANSCRIPT
The next page of the survey says:
> We’ll ask you about a variety of topics in this survey.
> In a moment, we’d like you to read a short news article. Please indicate which of the following you would be interested in reading about. You may not receive your preferred topic.
> 1.        The stock market
> 2.        Wages and income
> 3.        Unemployment
>
> The first option (the stock market) is selected.

The next page of the survey says:
> Please read the short new article below.
>
> WASHINGTON --- Stocks were up last week, lending further evidence that the markets are headed for even stronger gains. The Dow Jones rallied substantially in the past quarter, as investors displayed a growing appetite for risk and a striking lack of concern about any reversal of fortune. ``Today's report continues to signal a solid investing environment that is likely to maintain its bullishness," said Omair Sharif, senior economist at RBS in Stamford, Connecticut.

The next page of the survey says:
> Would you say that in the past year, BLUE-COLLAR WORKERS have been better off, the same, or worse off?
> Please choose a number from 1 (Much worse) to 5 (Much better)

CONDITION B TRANSCRIPT
The next page of the survey says:
> We’ll ask you about a variety of topics in this survey.
> In a moment, we’d like you to read a short news article. Please indicate which of the following you would be interested in reading about. You may not receive your preferred topic.
> 1.        The stock market
> 2.        Wages and income
> 3.        Unemployment
>
> The first option (the stock market) is selected.

The next page of the survey says:
> Please read the short new article below.
>
> NEW YORK--- Verizon once again has bragging rights in the wireless industry.
> The carrier took top honors in a new ranking of nationwide network performance by market research firm RootMetrics. Verizon ranked number one for reliability, speed, call and data performance, with AT&T a close second in all those categories. Those two firms held a significant advantage over third-ranked T-Mobile and fourth-ranked Sprint. RootMetrics conducted the study through the use of national surveys.

The next page of the survey says:
> Would you say that in the past year, BLUE-COLLAR WORKERS have been better off, the same, or worse off?
> Please choose a number from 1 (Much worse) to 5 (Much better)

Estimate the probability that the true population mean outcome under Condition A is strictly greater than the true population mean outcome under Condition B. Equivalently, estimate P(mean_A - mean_B > 0).

Provide your reasoning first. Then output valid JSON on a new line in exactly this structure:
{"p_positive": <number>}

The number must be between 0 and 1.
\end{Verbatim}

\subsubsection{\texttt{experiment\_priority\_no\_threshold}}
\begin{Verbatim}[breaklines=true,fontsize=\footnotesize]
[Study setting, outcome description, and Condition A/B transcripts identical to the first scientific-forecasting prompt; omitted.]

Suppose researchers have enough resources to carry forward only one of these two conditions into a new real experiment. The experiment would have the same cost, sample size, and measurement quality for either condition. The goal is to prioritize treatments expected to produce the largest positive effect on the stated outcome.

Which condition should the researchers test?

Provide your reasoning first. Then output exactly one multiple-choice answer on a new line in this exact format:
Choice: <one option exactly as written below>

Options:
A. Prioritize an experiment on Condition A
B. Prioritize an experiment on Condition B
\end{Verbatim}

\subsubsection{Threshold decision template}
\noindent\textit{Values used:} \verb|{threshold}| = 5\%, 10\%, 20\%, or 33.3\%.
\begin{Verbatim}[breaklines=true,fontsize=\footnotesize]
[Study setting, outcome description, and Condition A/B transcripts identical to the first scientific-forecasting prompt; omitted.]

Prioritize an experiment on Condition A if the probability that its true population mean outcome is strictly greater than Condition B's is at least {threshold}. Otherwise, prioritize an experiment on Condition B.

Provide your reasoning first. Then output exactly one multiple-choice answer on a new line in this exact format:
Choice: <one option exactly as written below>

Options:
A. Prioritize an experiment on Condition A
B. Prioritize an experiment on Condition B
\end{Verbatim}

\subsubsection{Point-payoff bet template}
\noindent\textit{Values used:} \verb|{gain}| = 19, 9, 4, or 2 points; the loss was always 1 point.
\begin{Verbatim}[breaklines=true,fontsize=\footnotesize]
[Study setting, outcome description, and Condition A/B transcripts identical to the first scientific-forecasting prompt; omitted.]

Consider the following bet:

- If the bet is accepted and the true population mean outcome under Condition A is strictly greater than under Condition B, the net gain is {gain} points.
- If the bet is accepted and the true population mean outcome under Condition A is not strictly greater than under Condition B, the net loss is 1 point.
- If the bet is declined, the payoff is 0 points regardless of which condition has the higher true mean outcome.

Based on the study information, would you recommend accepting this bet?

Provide your reasoning first. Then output exactly one multiple-choice answer on a new line in this exact format:
Choice: <one option exactly as written below>

Options:
A. Accept the bet
B. Decline the bet
\end{Verbatim}

\subsubsection{Asymmetric-cost decision template}
\noindent\textit{Values used:} \verb|{cost_ratio}| = 19, 9, 4, or 2.
\begin{Verbatim}[breaklines=true,fontsize=\footnotesize]
[Study setting, outcome description, and Condition A/B transcripts identical to the first scientific-forecasting prompt; omitted.]

Suppose researchers have enough resources to prioritize only one condition for testing in a new real experiment. Assume that the cost of a false negative—prioritizing Condition B when Condition A actually has the strictly higher true population mean—is {cost_ratio} times the cost of a false positive—prioritizing Condition A when Condition A does not have the strictly higher true population mean. Correct decisions have zero cost.

Which experiment should be prioritized?

Provide your reasoning first. Then output exactly one multiple-choice answer on a new line in this exact format:
Choice: <one option exactly as written below>

Options:
A. Prioritize an experiment on Condition A
B. Prioritize an experiment on Condition B
\end{Verbatim}

\subsubsection{\texttt{likert}}
\begin{Verbatim}[breaklines=true,fontsize=\footnotesize]
[Study setting, outcome description, and Condition A/B transcripts identical to the first scientific-forecasting prompt; omitted.]

How likely is it that the true population mean outcome under Condition A is strictly greater than the true population mean outcome under Condition B? Equivalently, how likely is it that mean_A - mean_B > 0?

Provide your reasoning first. Then output exactly one multiple-choice answer on a new line in this exact format:
Choice: <one option exactly as written below>

Options:
A. Very unlikely
B. Somewhat unlikely
C. Neither likely nor unlikely
D. Somewhat likely
E. Very likely
\end{Verbatim}

\subsubsection{\texttt{probability\_bin\_quintile}}
\begin{Verbatim}[breaklines=true,fontsize=\footnotesize]
[Study setting, outcome description, and Condition A/B transcripts identical to the first scientific-forecasting prompt; omitted.]

Which bin contains the probability that the true population mean outcome under Condition A is strictly greater than the true population mean outcome under Condition B? Equivalently, which bin contains P(mean_A - mean_B > 0)?

Provide your reasoning first. Then output exactly one multiple-choice answer on a new line in this exact format:
Choice: <one option exactly as written below>

Options:
A. Less than 20%
B. 20% to less than 40%
C. 40% to less than 60%
D. 60% to less than 80%
E. 80% or higher
\end{Verbatim}

\clearpage

\end{document}